\documentclass[sigconf]{acmart}

\renewcommand\footnotetextcopyrightpermission[1]{} 
\usepackage{url}
\usepackage{graphicx}
\usepackage{amsmath}
\usepackage{amsthm}
\usepackage{amssymb}
\usepackage{algorithm}
\usepackage{algorithmic}
\usepackage{bm}
\usepackage{subfigure}
\usepackage{caption}
\usepackage{enumitem}
\usepackage{natbib}
\usepackage{multirow}
\usepackage{color}
\usepackage{array}
\usepackage{nicefrac}

\newtheorem{theorem}{Theorem}[section]

\newtheorem{proposition}[theorem]{Proposition}
\newtheorem{definition}{Definition}[section]

\newcommand{\NM}[2]{\|  #1 \|_{#2}}
\newcommand{\Tr}[1]{\text{Tr}(#1)}
\newcommand{\Px}[2]{\text{\normalfont prox}_{#1}( #2 ) }
\newcommand{\Rl}[0]{\mathbb{R}}

\usepackage{array}
\newcolumntype{L}[1]{>{\raggedright\let\newline\\\arraybackslash\hspace{0pt}}m{#1}}
\newcolumntype{C}[1]{>{\centering\let\newline  \\\arraybackslash\hspace{0pt}}m{#1}}
\newcolumntype{R}[1]{>{\raggedleft\let\newline \\\arraybackslash\hspace{0pt}}m{#1}}

\newcommand{\w}{\bm{w}}

\AtBeginDocument{%
  \providecommand\BibTeX{{%
    \normalfont B\kern-0.5em{\scshape i\kern-0.25em b}\kern-0.8em\TeX}}}

\acmConference[---]{---}{---}{---}

\begin{document}


\title{Efficient Neural Interaction Function Search \\ for Collaborative Filtering}

\author{Quanming Yao}
\email{yaoquanming@4paradigm.com}
\authornote{Q. Yao is the corresponding author;
	X. Chen and Q. Yao contributed equally to this research;
	and the work is performed when X. Chen was an intern in 4Paradigm.}
\affiliation{%
	\institution{4Paradigm Inc. (Hong Kong)}
}

\author{Xiangning Chen}
\email{xiangning@cs.ucla.edu}
\affiliation{%
	\institution{Compute Science, UCLA}
}

\author{James T. Kwok}
\email{jamesk@cse.ust.hk}
\affiliation{%
	\institution{Dept. of Computer Science and Engineering, HKUST}
}

\author{Yong Li}
\email{liyong07@tsinghua.edu.cn}
\affiliation{%
	\institution{Dept. of Electronic Engineering  Tsinghua University}
}

\author{Cho-Jui Hsieh}
\email{chohsieh@cs.ucla.edu}
\affiliation{%
	\institution{Compute Science, UCLA}
}

\renewcommand{\shortauthors}{}

\begin{abstract}
In collaborative filtering (CF),
interaction function (IFC) play the
important role of 
capturing interactions among items and users.
The most popular IFC 
is the inner product, which has been
successfully used in low-rank matrix factorization.
However,
interactions 
in real-world applications
can be highly complex.
Thus,
other operations (such as plus and concatenation),
which may potentially offer better performance,
have been proposed.
Nevertheless,
it is still hard 
for existing IFCs 
to have consistently good performance across different application scenarios. 
Motivated by the recent success of automated machine learning (AutoML),
we propose 
in this paper
the search for simple neural interaction functions (SIF) in CF.
By examining and generalizing existing CF approaches,
an expressive SIF search space
is designed
and represented as a structured multi-layer perceptron.
We propose an one-shot search algorithm
that simultaneously updates both the architecture and learning parameters.
Experimental results demonstrate that
the proposed method 
can be much more efficient than popular AutoML approaches, 
can obtain much better prediction performance than state-of-the-art CF approaches,
and can discover distinct IFCs for different data sets and tasks.\footnote{Code is available at \url{https://github.com/xiangning-chen/SIF}.}
\end{abstract}

\ccsdesc[500]{Information systems~Collaborative filtering}
\ccsdesc[500]{Computing methodologies~Optimization algorithms}
\ccsdesc[500]{Computing methodologies~Continuous space search}
\ccsdesc[500]{Computing methodologies~Machine learning algorithms}

\keywords{Collaborative Filtering, automated machine learning, recommeder system, neural architecture search}

\maketitle

\begin{table*}[ht]
\centering
\caption{Popular human-designed interaction functions (IFC) for CF,
	where $\bm{H}$ is a parameter to be trained.
	SIF searches a proper IFC from the validation set (i.e., by AutoML),
	while others are all designed by experts.}
\label{tab:operations}
\vspace{-10px}
\renewcommand\arraystretch{1.20}
\begin{tabular}{c | C{90px} | C{60px} | C{50px} | C{45px} | C{70px}}
	\hline
	               &                                        IFC                                        &         operation         &     space     &  predict time  &                             recent examples                             \\ \hline
	               &                  $\left\langle \bm{u}_i, \bm{v}_j \right\rangle$                  &    {\sf inner product}    & $O((m + n)k)$ &     $O(k)$     & MF \citep{Koren2008FactorizationMT}, FM \citep{rendle2012factorization} \\ \cline{2-6}
	               &                               $\bm{u}_i - \bm{v}_j$                               & {\sf plus} ({\sf minus})  & $O((m + n)k)$ &     $O(k)$     &                   CML \citep{hsieh2017collaborative}                    \\ \cline{2-6}
	               &                     $\max\left( \bm{u}_i, \bm{v}_j \right) $                      &   {\sf max}, {\sf min}    & $O((m + n)k)$ &     $O(k)$     &                   ConvMF \citep{kim2016convolutional}                   \\ \cline{2-6}
	human-designed &                   $\sigma\left( [\bm{u}_i; \bm{v}_j] \right) $                    &       {\sf concat}        & $O((m + n)k)$ &     $O(k)$     &                    Deep\&Wide \citep{cheng2016wide}                     \\ \cline{2-6}
	               & $\sigma\left( \bm{u}_i\odot\bm{v}_j+\bm{H}\left[\bm{u}_i;\bm{v}_j\right] \right)$ & {\sf multi}, {\sf concat} & $O((m + n)k)$ &    $O(k^2)$    &                          NCF \citep{NeuMF2017}                          \\ \cline{2-6}
	               &                               $\bm{u}_i * \bm{v}_j$                               &        {\sf conv}         & $O((m + n)k)$ & $O(k \log(k))$ &                   ConvMF \citep{kim2016convolutional}                   \\ \cline{2-6}
	               &                            $\bm{u}_i \otimes \bm{v}_j$                            &    {\sf outer product}    & $O((m + n)k)$ &    $O(k^2)$    &                       ConvNCF \citep{he2018outer}                       \\ \hline\hline
	    AutoML     &                                  SIF (proposed)                                   &         searched          & $O((m + n)k)$ &     $O(k)$     &                                 ------                                  \\ \hline
\end{tabular}
\end{table*}

\section{Introduction}
\label{sec:intro}

Collaborative filtering (CF) \citep{herlocker1999algorithmic,su2009survey}
is an important topic in machine learning and data mining.
By capturing interactions among the rows and columns
in a data matrix,
CF predicts the missing entries 
based on the observed elements.
The most famous CF application 
is the recommender system \citep{Koren2008FactorizationMT}.
The ratings in such systems can be arranged as a data matrix,
in whch the rows correspond to users, the columns are items, and the entries are collected ratings.
Since users usually only interact with a few items, there are lots of missing entries in the rating matrix.
The task is to estimate users' ratings on items that they have not yet explored.
Due to the good empirical performance,
CF also have been used in various other applications.
Examples include
image inpainting in computer vision \citep{ji2010robust},
link prediction in social networks \citep{kim2011network}
and topic modeling for text analysis \cite{wang2011collaborative}.
More recently,
CF is also extended to tensor data (i.e., higher-order matrices) \cite{Kolda2009}
for the incorporation of side information
(such as extra features \cite{karatzoglou2010multiverse} and time
\citep{lei2009analysis}).

In the last decade,
low-rank matrix factorization \citep{Koren2008FactorizationMT,mnih2008probabilistic}
has been the most popular approach to CF.
It can be formulated as the following optimization problem:
\begin{align}
	\min_{\bm{U}, \bm{V}}
	\sum\nolimits_{(i,j)\in\Omega}
	\ell \left( \bm{u}_i^{\top} \bm{v}_j, \bm{O}_{ij} \right)
	+ \frac{\lambda}{2} \NM{\bm{U}}{F}^2
	+ \frac{\lambda}{2} \NM{\bm{V}}{F}^2,
	\label{eq:matcomp}
\end{align}
where $\ell$ is a loss function.
The observed elements are indicated by $\Omega$
with values given by the corresponding positions in matrix $\bm{O} \in \Rl^{m \times n}$,
$\lambda \ge 0$ is a hyper-parameter,
and $\bm{u}_i,\bm{v}_j \in \Rl^{k}$ are 
embedding vectors for 
user 
$i$
and 
item
$j$,
respectively.
Note that \eqref{eq:matcomp} captures interactions between user
$\bm{u}_i$ 
and item
$\bm{v}_j$
by the \textit{inner product}.
This achieves good empirical performance,
enjoys sound statistical guarantees \citep{candes2009exact,recht2010guaranteed}
(e.g., the data matrix can be exactly recovered when 
$\bm{O}$ 
satisfies
certain incoherence conditions
and the missing entires follow some distributions), 
and fast training \citep{mnih2008probabilistic,gemulla2011large}
(e.g., can be trained end-to-end by stochastic optimization).

While the inner product has many benefits,
it may not yield the best performance for various CF tasks
due to the complex nature of
user-item 
interactions.
For example,
if  $i$th and $j$th users like the $k$th item very much,
their embeddings should be close to each other
(i.e., $\NM{\bm{u}_i - \bm{u}_j}{2}$ is small).
This motivates the usage of the \textit{plus} operation \citep{NeuMF2017,hsieh2017collaborative},
as the triangle inequality ensures $\NM{\bm{u}_i - \bm{u}_j}{2} \le \NM{\bm{u}_i - \bm{v}_k}{2} + \NM{\bm{u}_j - \bm{v}_k}{2}$.
%
%
%
Other operations (such as concatenation and convolution)
have also outperformed the inner product on many CF tasks
\citep{rendle2012factorization,kim2016convolutional,he2018outer}.
Due to the success of deep networks \citep{goodfellow2016deep}, the
multi-layer perceptron (MLP) is recently
used 
as the interaction function (IFC) 
in CF 
\citep{cheng2016wide,NeuMF2017,Xue2017DeepMF},
and achieves good performance.
However,
choosing and designing an IFC is not easy, as 
it should depend on the data set and task.
Using one simple operation
may not be expressive enough to ensure good performance.
On the other hand,
directly using a MLP leads to 
the difficult and time-consuming task
of 
architecture selection
\citep{zoph2017neural,baker2017designing,zhang2019autokge}.
Thus,
it is hard to have a good
IFC across different tasks and data sets \cite{dacrema2019we}.

In this paper,
motivated by the success of 
automated machine learning (AutoML) \citep{automl_book,yao2018taking},
we consider formulating the search for interaction functions (SIF) as an AutoML problem.
Inspired by observations on existing IFCs,
we 
first
generalize the CF objective 
and define the SIF problem.
These observations also help to identify a domain-specific and expressive search space,
which not only includes many human-designed IFCs,
but also covers new ones not yet explored in the literature.
We further represent the SIF problem, armed with the designed search space, as a structured MLP.
This enables us
to derive an efficient  search algorithm
based on one-shot neural architecture search \citep{liu2018darts,xie2018snas,yao2019differentiable}. 
The 
algorithm
can jointly train the embedding vectors and search IFCs
in a stochastic end-to-end manner.
We further extend the proposed SIF,
including both the search space and one-shot search algorithm, 
to handle tensor data.
Finally,
we perform experiments on CF tasks with both matrix data 
(i.e., MovieLens data)
and tensor data (i.e., Youtube data).
The contributions of this paper are highlighted as follows:
\begin{itemize}[leftmargin = 9px]
\item The design
of interaction functions is a key issue in CF,
and is also a very hard problem due to varieties in the data sets and tasks (Table~\ref{tab:operations}).
We generalize the objective of CF,
and formulate the design of IFCs as an AutoML problem.
This is also the first work which
introduces AutoML techniques to CF.

\item 
By analyzing the formulations of existing IFCs,
we design an expressive but compact search space for the AutoML problem.
This covers previous IFCs given by experts as special cases, and
also allows generating novel IFCs that are new to the literature.
Besides,
such a search space can be easily extended
to handle CF problems on tensor data.

\item We 
propose an one-shot search algorithm
to efficiently optimize the AutoML problem.
This algorithm can jointly update
the architecture of IFCs (searched on the validation set) 
and the embedding vectors (optimized on the training set).

\item 
Empirical results demonstrate that,
the proposed algorithm can find better IFCs 
than existing AutoML approaches,
and is also much more efficient.
Compared with the human-designed CF methods,
the proposed algorithm can achieve much better performance,
while the computation cost is slightly higher than that from fine-tuning by experts.
To shed light on the design of IFCs,
we also 
perform a case study 
to show 
why better IFCs can be found by the proposed method.
\end{itemize}

\noindent
\textbf{Notations.}
Vectors are denoted by lowercase boldface,
and matrices by uppercase boldface.
For two vectors $\bm{x}$ and $\bm{y}$,
$\left\langle \bm{x}, \bm{y} \right\rangle$ is the inner product,
$\bm{x} \odot \bm{y}$ is the element-wise product,
$\bm{x} \otimes \bm{y}$ is the outer product,
$[\bm{x}; \bm{y}]$ concatenates (denoted ``concat'') two vectors to a longer one,
and $\bm{x} * \bm{y}$ is the convolution (denoted ``conv'').
$\Tr{\bm{X}}$ is the trace of a square matrix 
$\bm{X}$, and
$\NM{\bm{X}}{F}$ is the Frobenius norm.
$\NM{\bm{x}}{2}$ is the $\ell_2$-norm of a vector
$\bm{x}$, and
$\NM{\bm{x}}{0}$ counts its number of nonzero elements.
The proximal step \cite{parikh2013proximal} associated with a function $g$ is defined
as 
$\Px{g}{\bm{z}}
= \arg\min_{\bm{x}} 
\frac{1}{2}\NM{\bm{z} - \bm{x}}{2}^2 + g(\bm{x})$.
Let $\mathcal{S}$ be a constraint and $\mathbb{I}(\cdot)$ be the indicator function,
i.e., 
if $\bm{x} \in \mathcal{S}$ then $\mathbb{I}(\mathcal{S}) = 0$ and $\infty$ otherwise,
then 
%
$\Px{\mathbb{I}(\mathcal{S})}{\bm{z}}
= \arg\min_{\bm{x}} 
\frac{1}{2}\NM{\bm{z} - \bm{x}}{2}^2$
s.t. 
$\bm{x} \in \mathcal{S}$
is also the projection operator,
which maps $\bm{z}$ on $\mathcal{S}$.

\section{Related Works}
\label{sec:review}

\subsection{Interaction Functions (IFCs)}
\label{sec:lr:review}

%

As discussed in Section~\ref{sec:intro},
the IFC is key to CF.
Recently, many CF models with different ICFs have been proposed.
Examples include the 
factorization machine (FM) \citep{rendle2012factorization},
collaborative metric learning (CML) \citep{hsieh2017collaborative},
convolutional matrix factorization (ConvMF) \citep{kim2016convolutional},
Deep \& Wide \citep{cheng2016wide},
neural collaborative filtering (NCF) \citep{NeuMF2017},
and convolutional neural collaborative filtering (ConvNCF) \citep{he2018outer}.
As can be seen
from Table~\ref{tab:operations},
many operations
other than the simple inner product have been used.
Moreover,
they have the same space complexity
(linear in $m$, $n$ and $k$),
but different time complexities.

The design of IFCs 
depends highly on the given data and task.
As shown in a recent benchmark paper \cite{dacrema2019we},
no single IFC consistently outperforms the others
across all CF tasks \citep{su2009survey,aggarwal2016recommender}.
Thus, 
it is important
to either select a proper IFC from a set of customized IFC's designed by humans,
or to design a new IFC which has not been visited in the literature.

\subsection{Automated Machine Learning (AutoML)}
\label{sec:rel:auto}

To ease the use and design of better machine learning models,
automated machine learning (AutoML) \citep{automl_book,yao2018taking} has become a recent hot topic. 
AutoML can be seen as a bi-level optimization problem,
as we need to search for hyper-parameters and design of the underlying machine learning model.

\subsubsection{General Principles}
In general,
the success of AutoML hinges on
two important questions:
\begin{itemize}[leftmargin = 9px]
\item 
\textit{What to search}:
In AutoML, the
choice of the \textit{search space} 
is extremely important.
On the one hand,
the space needs to be general enough,
meaning that it should include human wisdom as special cases.
On the other hand,
the space cannot be too general,
otherwise the cost of searching in such a space can be too expensive \citep{zoph2017learning,liu2018darts}.
For example,
early works on neural architecture search (NAS) use reinforcement learning (RL) to
search among all possible designs of a 
convolution neural network (CNN) \citep{baker2017designing,zoph2017neural}.
This takes more than one thousand GPU days to 
obtain an architecture with performance comparable to the human-designed ones.
Later,
the search space is partitioned into blocks \citep{zoph2017learning},
which helps reduce the cost of RL to several weeks.

\item
\textit{How to search efficiently}:
Once the search space is determined,
the \textit{search algorithm} 
then matters.
Unlike convex optimization,
there is no universal and efficient optimization for AutoML \citep{automl_book}.
We need to invent efficient algorithms to find good designs in the space.
Recently, gradient descent based algorithms are adapted for NAS \cite{yao2019differentiable,liu2018darts,xie2018snas},
allowing joint update of the architecture weights and learning parameters.
This further reduces the search cost to one GPU day. 
\end{itemize}

\subsubsection{One-Shot Architecture Search Algorithms}
\label{sec:rel:osas}

Recently,
one-shot architecture search \cite{bender2018understanding} methods
such as DARTS~\cite{liu2018darts} and SNAS~\cite{xie2018snas},
have become the most popular NAS methods for the efficient search of good architectures. 
These methods
construct a supernet,
which contains all possible architectures spanned 
by the selected operations,
and then jointly optimize the network weights and architectures' parameters by stochastic gradient descent.
The state-of-the-art is 
NASP \citep{yao2019differentiable} (Algorithm~\ref{alg:nasp}). 
Let $\bm{\alpha}=[a_k]\in \mathbb{R}^d$, with $a_k$ encoding the weight of the
$k$th operation, and
$\bm{X}$ be the parameter.
In NSAP,
the selected operation $\bar{\mathcal{O}}(\bm{X})$ 
is represented as
\begin{align}\label{eq:oa}
\bar{\mathcal{O}}(\bm{X}) =  \sum\nolimits_{k=1}^{d}a_k \mathcal{O}_k(\bm{X}),~\text{\;where\;} 
~\bm{\alpha} \in \mathcal{C}_1 \cap \mathcal{C}_2,
\end{align}
$\mathcal{O}_k(\cdot)$ is the $k$th operation in $\mathcal{O}$, 
\begin{align}
\mathcal{C}_1
= \left\lbrace \bm{\alpha} \,|\, \NM{\bm{\alpha}}{0} = 1 \right\rbrace
\quad\text{and}\quad
\mathcal{C}_2 = \left\lbrace \bm{\alpha} \,|\, 0 \le \alpha_k \le 1 \right\rbrace.
\label{eq:spcons}
\end{align}
The discrete constraint in \eqref{eq:oa} forces only one operation to be selected. 
The search problem is 
then 
formulated as 
\begin{align}
\min_{\bm{\alpha}} 
\bar{\mathcal{L}}\left(\w^*(\bm{\alpha}), \bm{\alpha} \right),
\text{\;s.t.\;}
\begin{cases}
\w^*(\bm{\alpha}) = \arg\min_{\w} \mathcal{L}\left( \w, \bm{\alpha} \right)
\\
\bm{\alpha} \in \mathcal{C}_1 \cap \mathcal{C}_2
\end{cases}
,
\label{eq:nasp}
\end{align}
where $\bar{\mathcal{L}}$ (resp. $\mathcal{L}$) is the loss on validation (resp. training) data. 
As NASP targets at selecting and updating only one operation, 
it maintains two architecture representations: a continuous $\bm{\alpha}$ to be updated by gradient descent (step~4
in Algorithm~\ref{alg:nasp})
and a discrete $\bar{\bm{\alpha}}$ (steps~3 and 5). 
Finally, the
network weight $\w$ is optimized
on the training data in step~6.
The following Proposition
shows
closed-form solutions to the proximal step in Algorithm~\ref{alg:nasp}.

\begin{algorithm}[ht]
\caption{Neural architecture search by proximal iterations (NASP) algorithm \citep{yao2019differentiable}.}
	\begin{algorithmic}[1]
		\STATE {\bf require}: A mixture of operations $\bar{\mathcal{O}}$ parametrized by \eqref{eq:oa}, parameter $\w$ and stepsize $\eta$;
		\WHILE{not converged}
		\STATE Obtain \textit{discrete} architecture representation  $\bar{\bm{\alpha}} = \Px{\mathcal{C}_1}{\bm{\alpha}}$;
		\STATE Update \textit{continuous} architecture representation
		\begin{align*}
		\bm{\alpha} = \Px{\mathcal{C}_2}{\bm{\alpha} - \nabla_{\bar{\bm{\alpha}}} \bar{\mathcal{L}}( \bar{\w}, \bar{\bm{\alpha}})},
		\end{align*}
		where $\bar{\w} \! = \! \w \! - \! \eta \nabla_{\w} \mathcal{L}(\w, \bar{\bm{\alpha}})$
		(is an approximation to $\w^*(\bar{\bm{\alpha}})$); 
		\STATE Obtain new \textit{discrete} architecture $\bar{\bm{\alpha}} = \Px{\mathcal{C}_1}{\bm{\alpha}}$;
		\STATE Update $\w$ using $\nabla_{\w} \mathcal{L}(\w, \bar{\bm{\alpha}})$ with $\bar{\bm{\alpha}}$;
		
		\ENDWHILE
		\RETURN Searched architecture $\bar{\bm{\alpha}}$.
	\end{algorithmic}
	\label{alg:nasp}
\end{algorithm}

\begin{proposition}[\cite{parikh2013proximal,yao2019differentiable}]
\label{pr:proxnasp}
Let $\bm{z} \in \mathbb{R}^d$.
(i) $\Px{\mathcal{C}_1}{\bm{z}} = z_i \bm{e}_i$,
where $i = \arg\max_{i = 1, \cdots, d} |z_i|$, and $\bm{e}_i$ is a one-hot vector
with only the $i$th element being 1.
(ii) $\Px{\mathcal{C}_2}{\bm{z}} = \tilde{\bm{z}}$,
where 
$\tilde{\bm{z}}_i = z_i$ if $z_i \in [0, 1]$,
$\tilde{\bm{z}}_i = 0$ if $z_i < 0$,
and 
$\tilde{\bm{z}}_i = 1$ otherwise.
\end{proposition}


\section{Proposed Method}

In Section~\ref{sec:review},
we have discussed the importance of IFCs,
and the difficulty of choosing or designing one for the given task and data.
Similar observations have also been made in designing neural networks,
which motivates NAS methods for deep networks 
\citep{baker2017designing,zoph2017neural,zoph2017learning,xie2018snas,liu2018darts,bender2018understanding,yao2019differentiable}.
Moreover,
NAS has been developed as a replacement of humans,
which can discover data- and task-dependent architectures with better performance.
Besides,
there is no absolute winner for IFCs \cite{dacrema2019we}, 
just like the deep network architecture also depends on data sets and tasks.
These inspire us to 
search for proper IFCs in CF by AutoML approaches.  

\subsection{Problem Definition}

First, 
we define the AutoML problem here and identify an expressive search space for IFCs, which
includes the various operations in Table~\ref{tab:operations}.
Inspired by generalized matrix factorization \citep{NeuMF2017,Xue2017DeepMF} and objective \eqref{eq:matcomp},
we propose the following generalized CF objective:
\begin{align}
\min
F(\bm{U}, \bm{V}, \bm{w})
\equiv 
& \sum\nolimits_{(i,j)\in\Omega}\ell( \bm{w}^{\top} f \left( \bm{u}_i, \bm{v}_j \right), \bm{O}_{ij} )
\label{eq:gencf}
\\
& + \frac{\lambda}{2}\NM{\bm{U}}{F}^2 + \frac{\lambda}{2}\NM{\bm{V}}{F}^2,
\;\text{s.t.}\;
\NM{\bm{w}}{2} \le 1,
\notag
\end{align}
where $f$ is the IFC
(which takes the user embedding vector
$\bm{u}_i$
and item embedding vector
$\bm{v}_j$
as input, and outputs a vector),
and $\bm{w}$ is a learning parameter.
Obviously, 
all the IFCs in Table~\ref{tab:operations} can be represented by 
using different $f$'s. 
The following Proposition shows  that
the constraint $\NM{\bm{w}}{2} \le 1$ 
is necessary
to ensure existence
of a solution.

\begin{proposition} \label{pr:consw}
	If $f$ is an operation shown in Table~\ref{tab:operations}
	and the $\ell_2$-constraint on $\bm{w}$ is removed,
	then $F$ in
	(\ref{eq:gencf}) has
	no nonzero optimal solution when $\lambda > 0$
	(proof is in Appendix~\ref{app:pr:consw}).
\end{proposition}



Based on above objective, 
we now define the AutoML problem,
i.e.,
searching interaction functions (SIF) for CF,
here.

\begin{definition}[AutoML problem] \label{def:automc}
Let $\mathcal{M}$ be a performance measure (the lower the better) defined on the validation
set $\bar{\Omega}$ (disjoint from $\Omega$),
and
$\mathcal{F}$ be a family of 
vector-valued
functions with two vector inputs.
The problem of searching for an interaction function (SIF)
is formulated as
\begin{align}
f^*
& =  \arg\min_{f \in \mathcal{F}} \sum\nolimits_{ (i,j)\in \bar{\Omega} }
\mathcal{M}( f(\bm{u}^*_i, \bm{v}^*_j)^{\top} \bm{w}^{*}, \bm{O}_{ij})
\label{eq:sif}
\\
& \;\text{s.t.} \;
\left[ \bm{U}^*, \bm{V}^*, \bm{w}^* \right] 
=  
\underset{\bm{U}, \bm{V}, \bm{w}}{\arg\min}
F(\bm{U}, \bm{V}, \bm{w}),
\notag
\end{align}
where $\bm{u}^*_i$ (resp.  $\bm{v}^*_j$) is the $i$th column of $\bm{U}^*$ (resp. $j$th column of $\bm{V}^*$).
\end{definition}

Similar to other AutoML problems (such as
auto-sklearn \citep{feurer2015efficient}, 
NAS \citep{baker2017designing,zoph2017neural} and AutoML in knowledge graph \cite{zhang2019autokge}),
SIF is a bi-level optimization problem \citep{colson2007overview}.
On the top level, 
a good architecture $f$ is searched based on the validation set.
On the lower level,
we find the model parameters using $F$ on the training set.
Due to the nature of bi-level optimization,
AutoML problems are 
difficult to solve
in general.
In the following,
we show how to design an expressive search space
(Section~\ref{sec:space}),
propose an efficient one-shot search algorithm (Section~\ref{sec:algorithm}),
and extend the proposed method to tensor data (Section~\ref{sec:tensor}).

\subsection{Designing a Search Space}
\label{sec:space}

Because of the powerful approximation capability of deep networks \citep{raghu2017expressive},
NCF \citep{NeuMF2017} and Deep\&Wide \citep{cheng2016wide} use a MLP as $f$.
SIF then becomes searching a suitable MLP 
from the family 
$\mathcal{F}$ based on the validation set (details are in Appendix~\ref{app:mlpdetails}),
where both the MLP architecture and weights are searched.
However,
a direct search of this MLP can be expensive and difficult,
since determining its architecture is already an extremely time-consuming problem
as observed in the NAS literature \citep{zoph2017learning,liu2018darts}.
Thus,
as in Section~\ref{sec:rel:auto},
it is preferable to use 
a simple but expressive search space
that exploits domain-specific knowledge from experts.

Notice that
Table~\ref{tab:operations} 
contains
operations that are
\begin{itemize}[leftmargin = 10px]
\item \textit{Micro (element-wise)}:
a possibly nonlinear function 
operating on individual elements,
and
\item \textit{Marco (vector-wise)}:
operators that operate on the whole 
input vector
(e.g.,
minus and multiplication).
\end{itemize}
Inspired by previous attempts that divide the 
NAS 
search space into micro and macro levels 
\citep{zoph2017learning,liu2018darts},
we propose to 
first search for a nonlinear transform on each single element, and
then combine these element-wise operations 
at the vector-level.
Specifically, let 
$\mathcal{O}$ be an operator selected from
\textsf{multi}, 
\textsf{plus},
\textsf{min},
\textsf{max},
\textsf{concat},
$g(\beta;\bm{x})\in {\mathbb R}$ be a simple nonlinear function
with input $\beta\in {\mathbb R}$ and hyper-parameter $\bm{x}$.
We construct a search space $\mathcal{F}$ for \eqref{eq:sif}, in which each $f$ is
\begin{align}
f(\bm{u}_i, \bm{v}_j)
= \mathcal{O} ( \dot{\bm{u}}_i, \dot{\bm{v}}_j ),
\label{eq:space}
\end{align} 
with $\left[ \dot{\bm{u}}_i \right]_l = g\left( \left[ \bm{u}_i \right]_l ; \bm{p} \right)$
and $[ \dot{\bm{v}}_j ]_l = g( [ \bm{v}_j ]_l ; \bm{q} )$
where $\left[ \bm{u}_i \right]_l$ (resp. $[ \bm{v}_j ]_l$) is the $l$th element of $\bm{u}_i$ (resp. $[ \bm{v}_j ]_l$),
and $\bm{p}$ (resp. $\bm{q}$) is the hyper-parameter of $g$ transforming  the
user (resp. item) embeddings.
Note that
we omit
the convolution and outer product
(vector-wise operations) 
from $\mathcal{O}$ 
in \eqref{eq:space},
as they need significantly more computational time
and have inferior performance than the rest (see Section~\ref{sec:exp:single}).
Besides,
we parameterize $g$ with a very small MLP with fixed architecture 
(single input, single output and five sigmoid hidden units)
for the element-wise level in \eqref{eq:space},
and the $\ell_2$-norms of the weights,
i.e., $\bm{p}$ and $\bm{q}$ in \eqref{eq:space},
are constrained to be smaller than or equal to 1.

This search space $\mathcal{F}$ meets the requirements for AutoML in Section~\ref{sec:rel:auto}.
First, 
as it involves an extra nonlinear transformation,
it contains operations that are more general than those designed by experts in
Table~\ref{tab:operations}.
This expressiveness leads to better performance than the human-designed models in the experiments (Section~\ref{ssec:exp:cf}).
Second,
the search space is much more constrained than that of a general MLP mentioned
above,
as we only need to select an operation for $\mathcal{O}$ and determine the weights for 
a small fixed MLP (see Section~\ref{ssec:exp:auto}).

\subsection{Efficient One-Shot Search Algorithm}
\label{sec:algorithm}

Usually,
AutoML problems 
require
full model training  and
are expensive to search.
In this section,
we propose an efficient algorithm,
which only approximately trains the models, and
to search the space in an end-to-end stochastic manner.
Our algorithm is motivated by 
the recent success of one-shot architecture search.

\subsubsection{Continuous Representation of the Space}
\label{sssec:rmlp}

Note that the search space in \eqref{eq:space} contains both discrete 
(i.e., choice of operations) and continuous variables 
(i.e., hyper-parameter $\bm{p}$ and $\bm{q}$ for nonlinear transformation).
This 
kind of search
is 
inefficient 
in general.
Motivated by differentiable search in NAS \citep{liu2018darts,xie2018snas},
we propose to relax the choices among operations
as a sparse vector in a continuous space.
Specifically,
we transform $f$ in \eqref{eq:space} as
\begin{align}
h_{\alpha} (\bm{u}_i, \bm{v}_j)
\equiv
\sum\nolimits_{m = 1}^{|\mathcal{O}|} 
\alpha_m
\left( 
\bm{w}_{m}^{\top}
\mathcal{O}_m 
(
\dot{\bm{u}}_i, 
\dot{\bm{v}}_j
)
\right) 
\quad\text{s.t.}\quad
\bm{\alpha} \in \mathcal{C},
\label{eq:relax}
\end{align}
where $\bm{\alpha} = [\alpha_m]$ and $\mathcal{C}$ (in \eqref{eq:spcons})
enforces that only one operation is selected.
Since operations may lead to different output sizes,
we associate each operation $m$ with its own $\bm{w}_m$.

\begin{figure}[ht]
	\centering
	\includegraphics[width=0.38\textwidth]{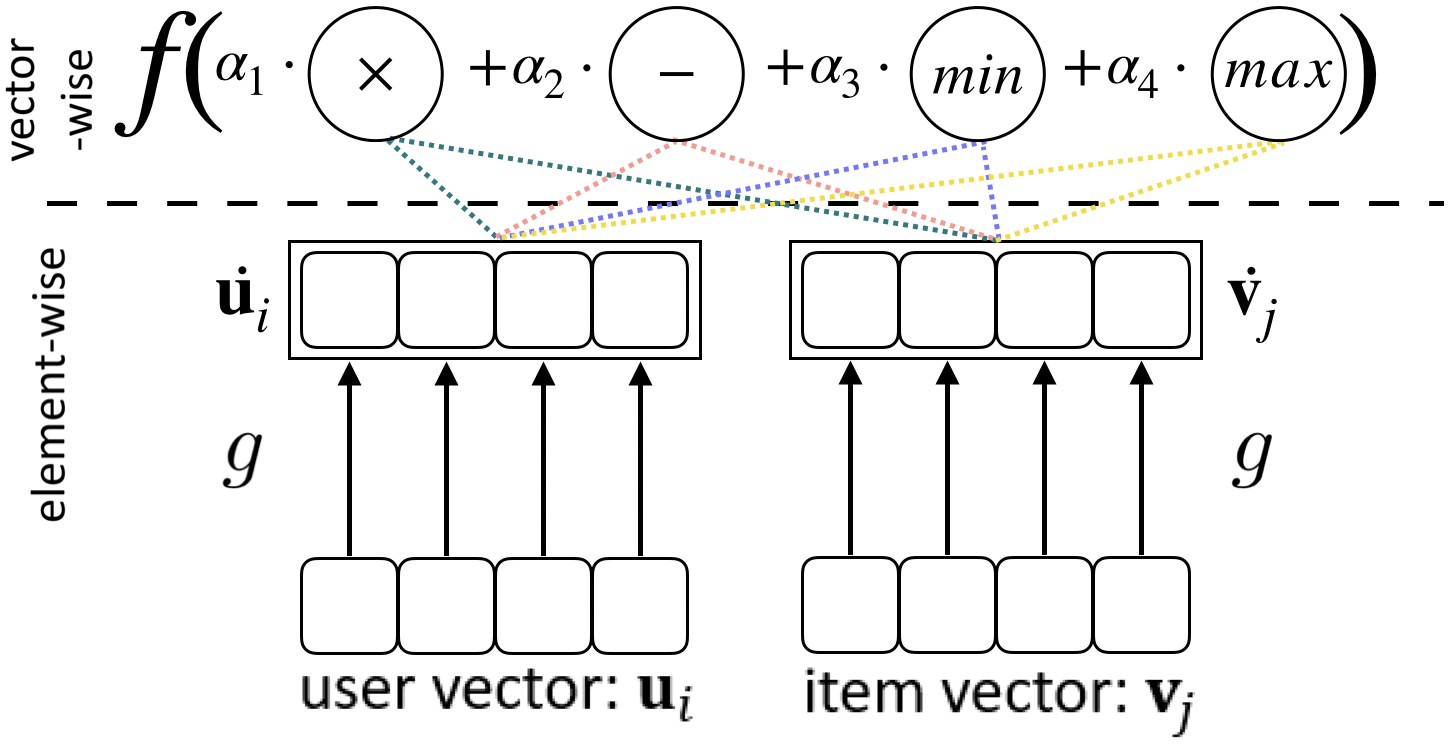}
	\vspace{-10px}
	\caption{Representing the search space as a structured MLP.
		Vector-wise: standard linear algebra operations;
		element-wise: simple non-linear transformation.}
	\label{fig:space}
\end{figure}

Let $\bm{T} = \{ \bm{U}, \bm{V}, \{ \bm{w}_{m} \} \}$ be the parameters to be
determined by the training data,
and $\bm{S} = \{ \bm{\alpha}, \bm{p}, \bm{q} \}$ be the hyper-parameters
to be determined by 
the validation set.
Combining $h_{\alpha}$ with \eqref{eq:sif},
we propose the
following objective:
\begin{eqnarray}
& \min_{ \bm{S} } &
H( \bm{S}, \bm{T} ) 
\equiv
\sum\nolimits_{ (i,j)\in \bar{\Omega} }
\mathcal{M}
( 
h_{\alpha} (\bm{u}^*_i, \bm{v}^*_j)^{\top} \bm{w}_{\alpha}^{*}, \bm{O}_{ij} 
)
\label{eq:pobj}
\\
& \text{s.t.} & 
\bm{\alpha} \in \mathcal{C}
\;\text{and}\;
\bm{T}^*
\equiv\{ \bm{U}^*, \bm{V}^*, \{ \bm{w}_m^* \} \}
= 
\arg\min_{\bm{T}}
F_{\alpha}(\bm{T}; \bm{S}),
\notag
\end{eqnarray}
where
$F_{\alpha}$ is
the training objective:
\begin{align*}
F_{\alpha}(\bm{T}; \bm{S})
\equiv
&
\sum\nolimits_{(i,j)\in\Omega}
\ell
( 
h_{\alpha} (\bm{u}_i, \bm{v}_j), \bm{O}_{ij}  
)
+ \frac{\lambda}{2}\NM{\bm{U}}{F}^2
+ \frac{\lambda}{2}\NM{\bm{V}}{F}^2,
\\
\;\text{s.t.}\;
&
\NM{\bm{w}_m}{2} \le 1
\;\text{for}\;
m = 1, \dots, |\mathcal{O}|.
\end{align*}
Moreover, 
the objective \eqref{eq:pobj} can be expressed as a structured MLP (Figure~\ref{fig:space}).
Compared with 
the general MLP mentioned in Section~\ref{sec:space},
the architecture of this structured MLP is fixed and its total number of parameters 
is very small.
After solving \eqref{eq:pobj},
we keep $\bm{p}$ and $\bm{q}$ for element-wise non-linear transformation,
and pick the operation which is indicated by the 
only nonzero position in the vector $\bm{\alpha}$ for vector-wise interaction.
The model 
is then re-trained 
to obtain the final user and item embedding
vectors
($\bm{U}$ and $\bm{V}$) and the corresponding $\bm{w}$ in \eqref{eq:gencf}.

\subsubsection{Optimization by One-Shot Architecture Search}
\label{sec:alg}

We present a stochastic algorithm (Algorithm~\ref{alg:sif}) to optimize 
the structured MLP in Figure~\ref{fig:space}.
The algorithm is inspired by NASP (Algorithm~\ref{alg:nasp}),
in which the relaxation of operations is defined in \eqref{eq:relax}.
Again,
we need to keep a discrete representation
of the architecture,
i.e., $\bar{\bm{\alpha}}$ at steps~3 and 8,
but optimize a continuous architecture,
i.e., $\bm{\alpha}$ at step~5.
The difference is that
we have extra continuous hyper-parameters $\bm{p}$ and $\bm{q}$ 
for element-wise nonlinear transformation here.
They can still be updated by proximal steps (step~6),
in which the closed-form solution is given by
$\Px{\mathbb{I}(\NM{\cdot}{2} \le 1)}{ \bm{z} } = \bm{z}/\NM{\bm{z}}{2}$
\cite{parikh2013proximal}.

%

\begin{algorithm}[ht]
	\caption{Searching Interaction Function (SIF) algorithm.}
	\begin{algorithmic}[1]
		\STATE Search space $\mathcal{F}$ represented by a structured MLP (Figure~\ref{fig:space});
		\WHILE{epoch $t = 1, \cdots, T$}
		\STATE Select one operation $\bar{\bm{\alpha}} = \Px{\mathcal{C}_1}{\bm{\alpha}}$; 
		\STATE \textit{sample a mini-batch from the validation data set};
		\STATE Update continuous $\bm{\alpha}$ for vector-wise operations
		\begin{align*}
		\bm{\alpha} = \Px{\mathcal{C}_2}{\bm{\alpha} - \eta \nabla_{\bar{\bm{\alpha}}} H (\bm{T}, \bm{S})};
		\end{align*}
		\vspace{-10px}
		\STATE Update element-wise transformation
		\begin{align*}
		\bm{p} & = \Px{ \mathbb{I}(\NM{\cdot}{2} \le 1) }{ \bm{p} - \eta \nabla_{\bm{p}} H (\bm{T}, \bm{S}) },
		\\
		\bm{q} & = \Px{ \mathbb{I}(\NM{\cdot}{2} \le 1) }{ \bm{q} - \eta \nabla_{\bm{q}} H (\bm{T}, \bm{S}) };
		\end{align*}
		\vspace{-10px}
		\STATE \textit{sample a mini-batch from the training data set};
		\STATE Obtain selected operation $\bar{\bm{\alpha}} = \Px{\mathcal{C}_1}{\bm{\alpha}}$;
		
		\STATE Update training parameters $\bm{T}$ with gradients on $F_{\alpha}$;
		\ENDWHILE
		\RETURN Searched interaction function (parameterized by $\bm{\alpha}$, $\bm{p}$ and $\bm{q}$,
		see \eqref{eq:space} and \eqref{eq:relax}).
	\end{algorithmic}
	\label{alg:sif}
\end{algorithm}

\subsection{Extension to Tensor Data}
\label{sec:tensor}

As mentioned in Section~\ref{sec:intro},
CF methods have also been used on tensor data.
For example, low-rank matrix factorization is extended to tensor factorization, in
which two decomposition formats, CP and Tucker \citep{Kolda2009}, have been popularly used.
These two methods are also based on the inner product.
Besides, the
factorization machine \citep{rendle2012factorization}
is also recently extended to data cubes \citep{blondel2016higher}.
These motivate us to extend the proposed SIF algorithm to tensor data.
In the sequel,
we focus on the third-order tensor.
Higher-order tensors can be handled in a similar way.

For tensors,
we need to maintain three embedded vectors,
$\bm{u}_i$, $\bm{v}_j$ and $\bm{s}_l$.
First,
we modify $f$ to take three vectors as input and output another vector.
Subsequently,
each candidate in search space \eqref{eq:space} 
becomes
$f = \mathcal{O}(\dot{\bm{u}}_i, \dot{\bm{v}}_j, \dot{\bm{s}}_l)$,
where $\dot{\bm{u}}_i$'s are obtained from element-wise MLP from $\bm{u}_i$
(and similarly for $\dot{\bm{v}}_j$ and $\dot{\bm{s}}_l$).
However, 
$\mathcal{O}$ is no longer a single operation,
as three vectors are involved.
$\mathcal{O}$ enumerates all possible combinations from basic operations in the matrix case.
For example,
if only $\max$ and $\odot$ are allowed,
then $\mathcal{O}$ contains
$\max(\bm{u}_i, \bm{v}_j) \odot \bm{s}_l$,
$\max(\max(\bm{u}_i, \bm{v}_j), \bm{s}_l)$,
$\bm{u}_i \odot \max(\bm{v}_j, \bm{s}_l)$
and $\bm{u}_i \odot \bm{v}_j \odot \bm{s}_l$.
With the above modifications,
it is easy to see that the space can still be represented by a structured MLP
similar to that in Figure~\ref{fig:space}.
Moreover, the proposed Algorithm~\ref{alg:sif} can still be applied (see Appendix~\ref{app:tensor}).
Note that the search space is much larger for tensor than matrix.

\section{Empirical Study}

\begin{figure*}[ht]
	\centering	
	\subfigure[MovieLens-100K.]
	{\includegraphics[width = 0.29\textwidth]{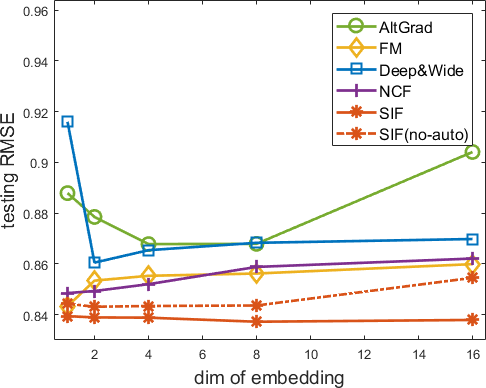}
		\label{fig:cf:100k}}
	\quad
	\subfigure[MovieLens-1M.]
	{\includegraphics[width = 0.29\textwidth]{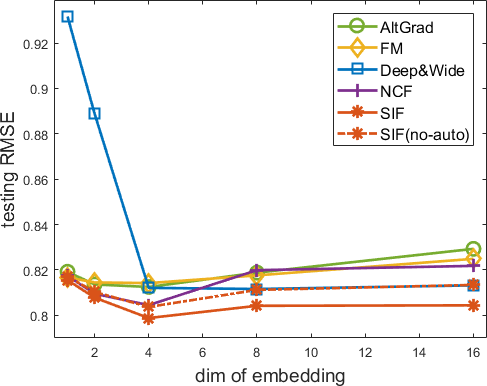}
		\label{fig:cf:1m}}
	\quad
	\subfigure[Youtube.]
	{\includegraphics[width = 0.29\textwidth]{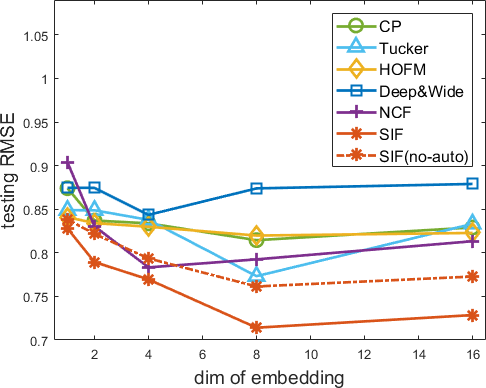}
		\label{fig:cf:ysub}}
	
	\vspace{-14px}
	\caption{Testing RMSEs of \textit{SIF} and other CF approaches
		with different embedding dimensions.}
	\label{fig:cf}
	\vspace{-5px}
\end{figure*}

\begin{figure*}[ht]
	\centering	
	\subfigure[MovieLens-100K.]
	{\includegraphics[width = 0.29\textwidth]{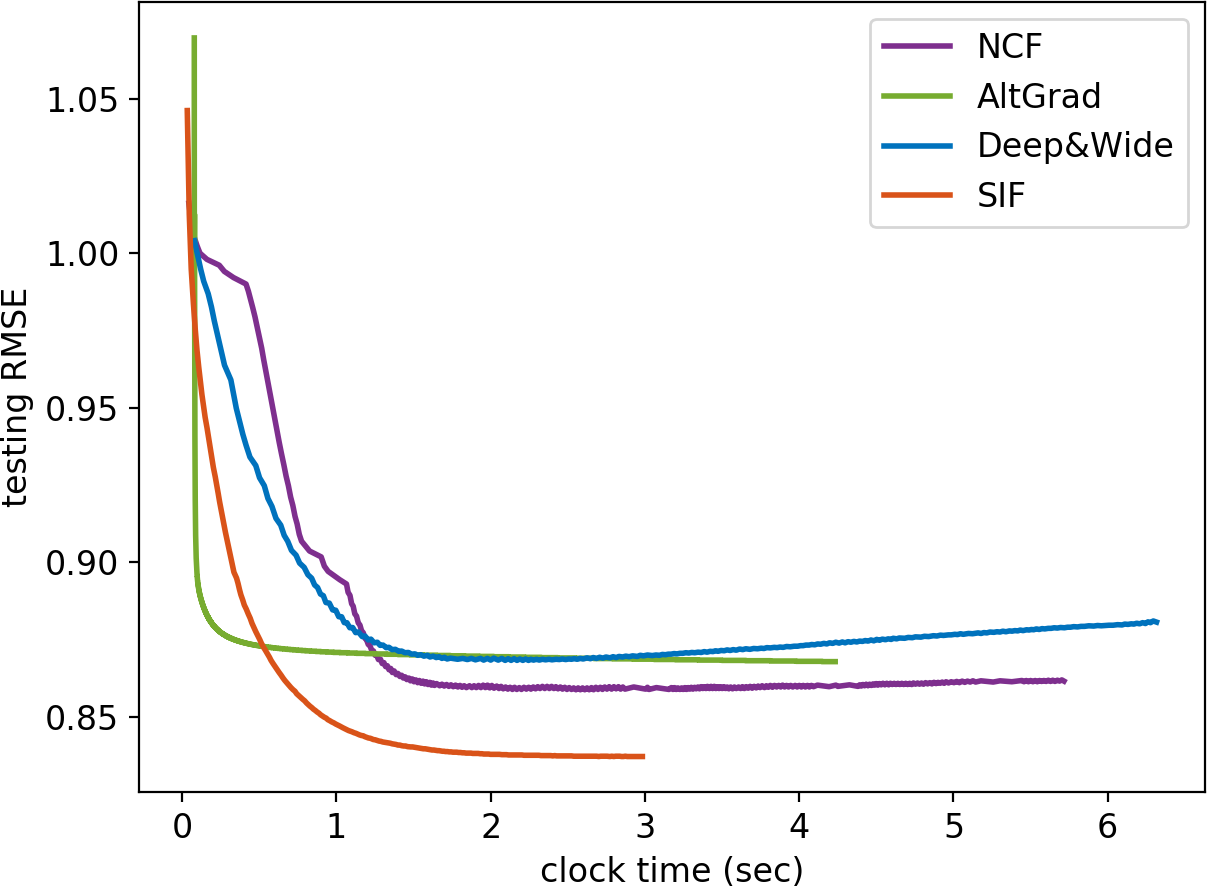}
		\label{fig:cfsp:100k}}
	\quad
	\subfigure[MovieLens-1M.]
	{\includegraphics[width = 0.29\textwidth]{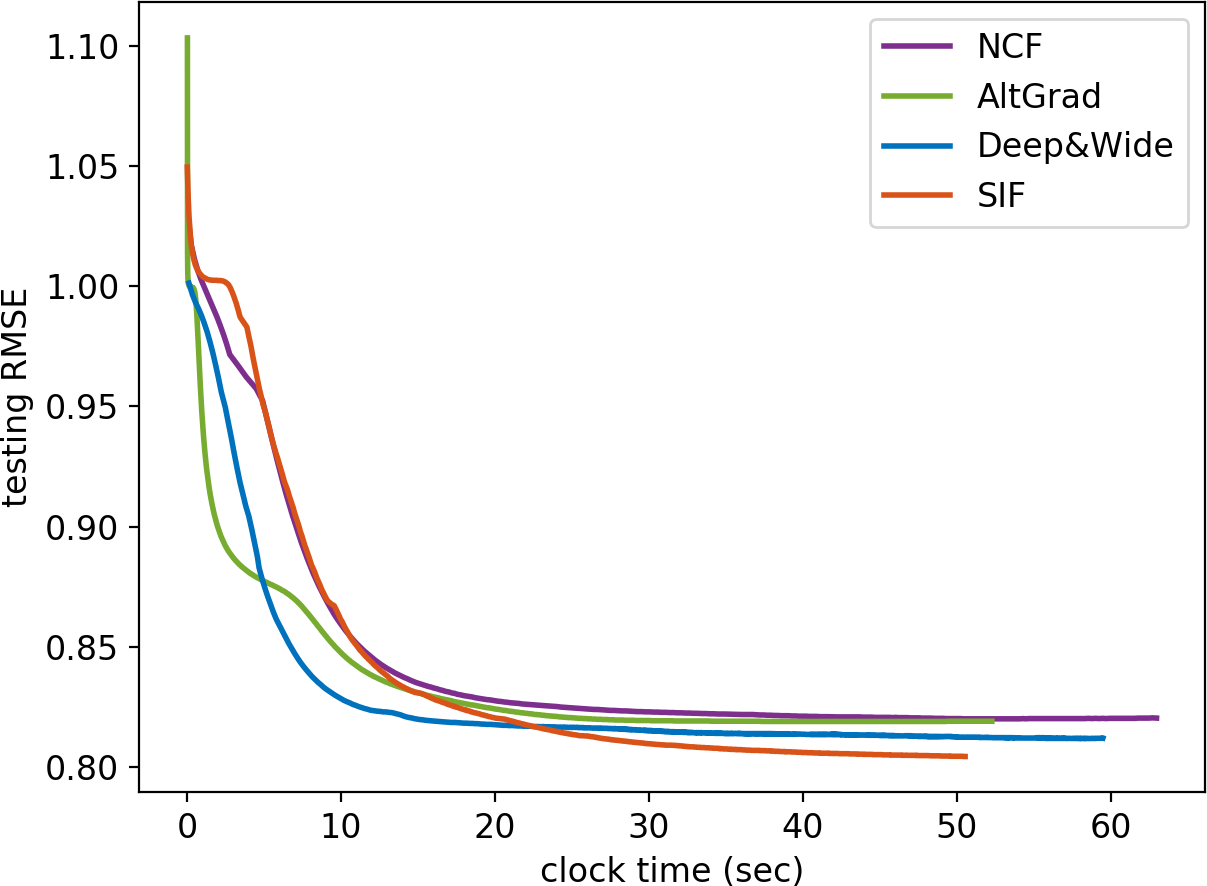}
		\label{fig:cfsp:1m}}
	\quad
	\subfigure[Youtube.]
	{\includegraphics[width = 0.29\textwidth]{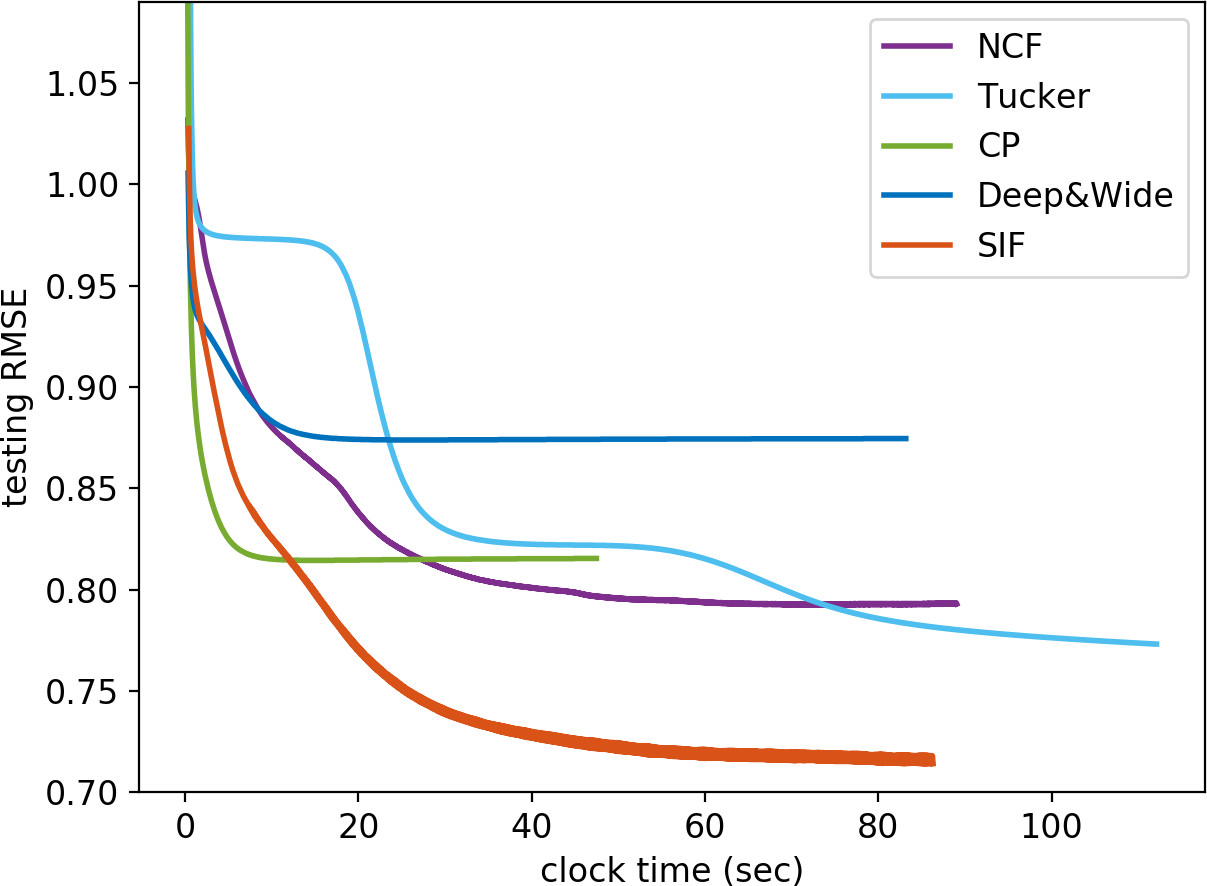}
		\label{fig:cfsp:ysub}}
	
	\vspace{-14px}
	\caption{Convergence of \textit{SIF} (with searched IFC)
		and other CF methods
		with an embedded dimensionality of $8$.
		Algorithms \textit{FM} and \textit{HOFM} are not shown as their codes do not support
		a callback to record testing performance.}
	\label{fig:cfsp}
	\vspace{-5px}
\end{figure*}

\subsection{Experimental Setup}

Two standard benchmark data sets 
(Table~\ref{tab:tensor_info}),
MovieLens
(matrix data) and 
Youtube
(tensor data),
are used in the experiments
\citep{mnih2008probabilistic,gemulla2011large,lei2009analysis}.
Following \citep{wang2015orthogonal,yao2018accelerated},
we uniformly and randomly select 50\% of the ratings for training,
25\% for validation and the rest for testing.
Note that since the size of the original Youtube dataset \cite{lei2009analysis} is very large 
(approximate 27 times the size of MovieLens-1M), 
we sample a subset of it to test the performance 
(approximately the size of MovieLens-1M). 
We sample rows with interactions larger than 20.

\begin{table}[ht]
	\centering
	\vspace{-5px}
	\caption{Data sets used in the experiments.}
	\vspace{-10px}
	\begin{tabular}{c c|c|c|c}
		\hline
		\multicolumn{2}{c|}{data set (matrix)} & \#users & \#items &  \#ratings  \\ \hline
		\multirow{2}{*}{MovieLens} & 100K      &  943  & 1,682  &  100,000  \\ \cline{2-5}
		                           & 1M        & 6,040  & 3,706  & 1,000,209 \\ \hline
	\end{tabular}
	
	\vspace{5px}
	
	\begin{tabular}{c|c|c|c|c}
		\hline
		data set (tensor) & \#rows & \#columns & \#depths & \#nonzeros \\ \hline
		     Youtube      & 600  & 14,340  &   5    & 1,076,946 \\ \hline
	\end{tabular}
	\label{tab:tensor_info}
	\vspace{-5px}
\end{table}

The task is to predict missing ratings given the training data.
We use the square loss for both $\mathcal{M}$ and $\ell$.
For performance evaluation,
we use (i)
the testing RMSE 
as in \citep{mnih2008probabilistic,gemulla2011large}:
$
\text{RMSE} = 
[ 
\frac{1}{| \tilde{\Omega} |} 
\sum\nolimits_{(i,j) \in \tilde{\Omega}} 
(\bm{w}^{\top} f(\bm{u}_i, \bm{v}_j)
- \bm{O}_{ij})^2
]^{\nicefrac{1}{2}},
$
where $f$ is the operation chosen by the algorithm,
and $\bm{w}$, $\bm{u}_i$'s and $\bm{v}_j$'s are parameters learned from
the training data;
and (ii) clock time (in seconds) as in \citep{baker2017designing,liu2018darts}.
Except for IFCs,
other hyper-parameters are all tuned with grid search on the validation set.
Specifically,
for all CF approaches, 
since the network architecture is already pre-defined,
we tune the learning rate $lr$ and regularization
coefficient $\lambda$ to obtain the best RMSE.
We use the Adagrad \cite{Duchi2010Adagrad} optimizer for gradient-based
updates.
In our experiments,
$lr$ is not sensitive, and
we simply fix it to a 
small value. Furthermore, we utilize grid search to obtain $\lambda$
from $\left[ 0,10^{-6},5 \times 10^{-6}, 10^{-5}, 5 \times 10^{-5}, 10^{-4} \right]$. 
For the AutoML approaches, we use the same $lr$ to search for the architecture, and
tune $\lambda$ using the same grid after the searched architecture is obtained.

\subsection{Comparison with State-of-the-Art CF Approaches}
\label{ssec:exp:cf}

In this section, we compare SIF with  
state-of-the-art 
CF approaches.
On the matrix data sets,
the following methods are compared:
(i) Alternating gradient descent (``\textit{AltGrad}'') \citep{Koren2008FactorizationMT}:
This is the most popular CF method,
which is based on matrix factorization (i.e., inner product operation).
Gradient descent 
is used
for optimization;
(ii) Factorization machine (``\textit{FM}'') \citep{rendle2012factorization}:
This extends linear regression with matrix factorization to capture second-order interactions among features; 
(iii)
\textit{Deep\&Wide} \citep{cheng2016wide}: This is a recent CF method.
It first embeds discrete features and then concatenates them for prediction;
(iv)
Neural collaborative filtering (``\textit{NCF}'') \citep{NeuMF2017}: 
This is	another recent CF method
which models the IFC by neural networks.

For tensor data, \textit{Deep\&Wide} and \textit{NCF} can be easily extended to tensor data.
Two types of popularly used low-rank factorization of tensor are used,
i.e.,  ``\textit{CP}'' and ``\textit{Tucker}'' \citep{Kolda2009},
and gradient descent is used for optimization;
``\textit{HOFM}'' \citep{blondel2016higher}: a fast variant of FM, which can capture high-order interactions among features.
Besides,
we also compare with
a variant of SIF (Algorithm~\ref{alg:sif}), denoted
\textit{SIF(no-auto)},
in which both the embedding parameter $\bm{T}$ and architecture parameter $\bm{S}$ are optimized using training data.
Details on the implementation of each CF method and discussion of the other CF approaches are in Appendix~\ref{app:details:cf}.
All codes are implemented in PyTorch, and run on a GPU cluster with a Titan-XP GPU. 

\subsubsection{Effectiveness}

Figure~\ref{fig:cf}
shows 
the testing RMSEs.
As the embedding dimension gets larger,
all methods gradually overfit and the testing RMSEs get higher.
\textit{SIF(no-auto)} is slightly better than the other CF approaches,
which demonstrates the expressiveness of the designed search space.
However,
it is worse than \textit{SIF}.
This shows that using the validation set can lead to better architectures.
Moreover, with the searched IFCs,
\textit{SIF} consistently obtains lower testing RMSEs than
the other CF approaches.

\subsubsection{Convergence}
If an IFC can
better 
capture the interactions among user and item embeddings,
it can also converge faster in terms of testing performance.
Thus, 
we show the training efficiency of the searched interactions
and human-designed CF methods in Figure~\ref{fig:cfsp}.
As can be seen,
the searched IFC can be more efficient, which again shows superiority of 
searching IFCs from data.

\begin{figure*}[ht]
	\centering
	\subfigure[MovieLens-100K.]
	{\includegraphics[width = 0.29\textwidth]{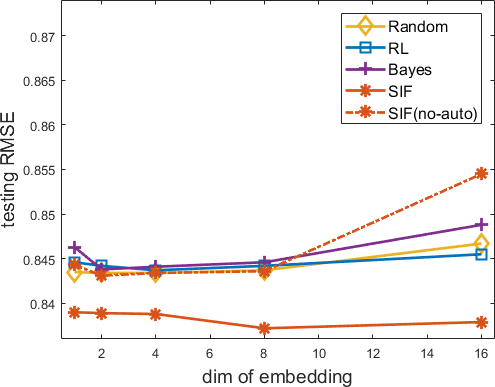}
		\label{fig:auto:100k}}	
	\quad
	\subfigure[MovieLens-1M.]
	{\includegraphics[width = 0.29\textwidth]{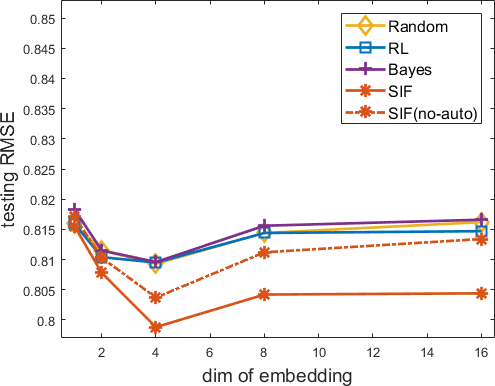}
		\label{fig:auto:1m}}
	\quad
	\subfigure[Youtube.]
	{\includegraphics[width = 0.285\textwidth]{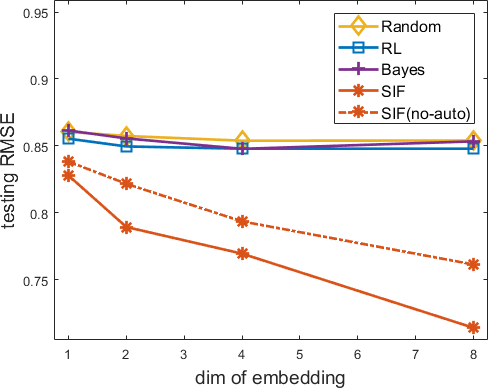}
		\label{fig:auto:ysub}}
	\vspace{-14px}
	\caption{Testing RMSEs of \textit{SIF} and the other AutoML approaches,
		with different embedding dimensions.
		\textit{Gen-approx} is 
		not run on Youtube, as it is
		slow and the 
		performance is inferior.}
	\label{fig:auto}
	\vspace{-5px}
\end{figure*}


\begin{figure*}[ht]
	\centering	
	\subfigure[MovieLens-100K.]
	{\includegraphics[width = 0.29\textwidth]{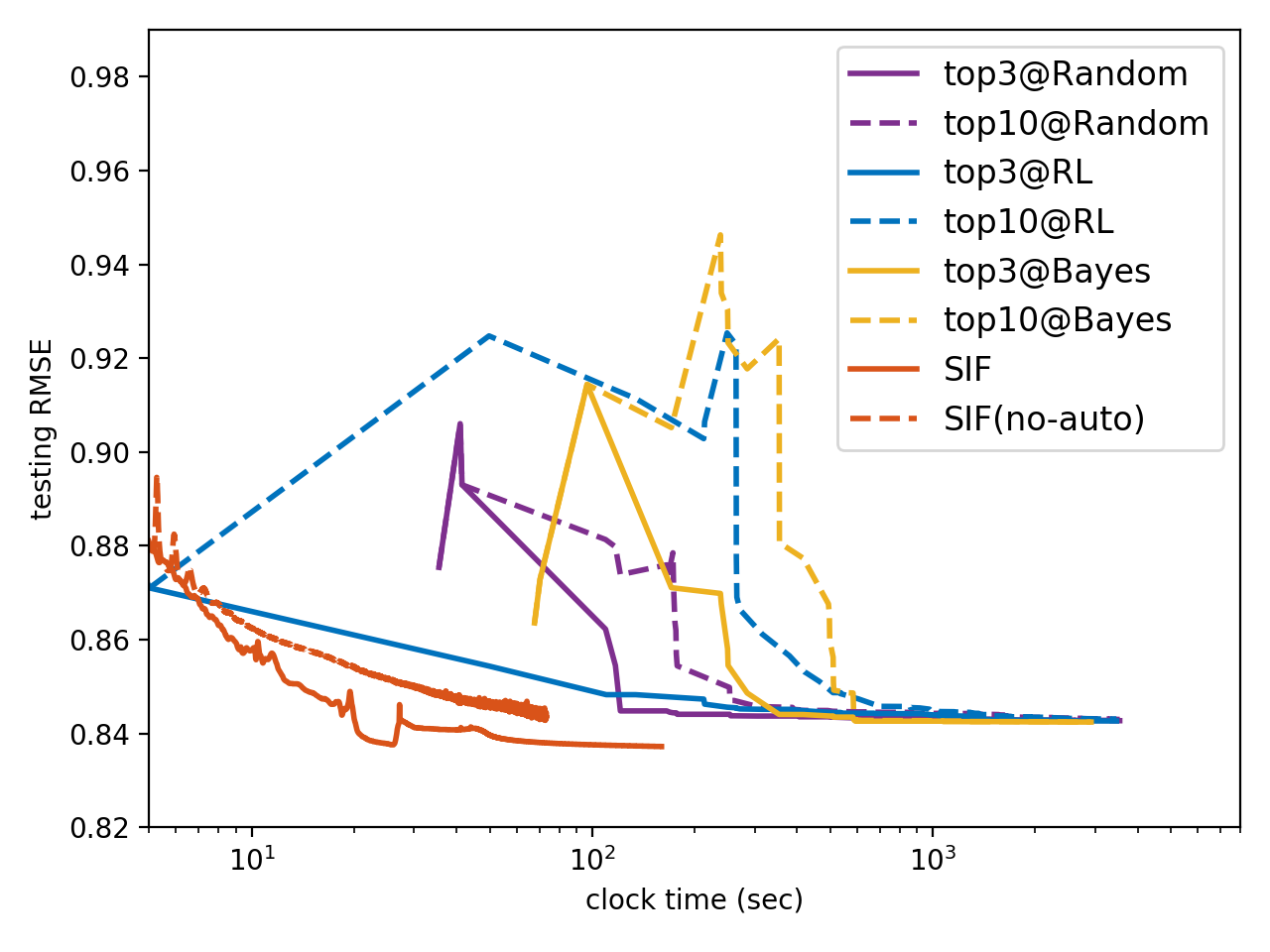}
		\label{fig:automc:100k}}
	\quad
	\subfigure[MovieLens-1M.]
	{\includegraphics[width = 0.29\textwidth]{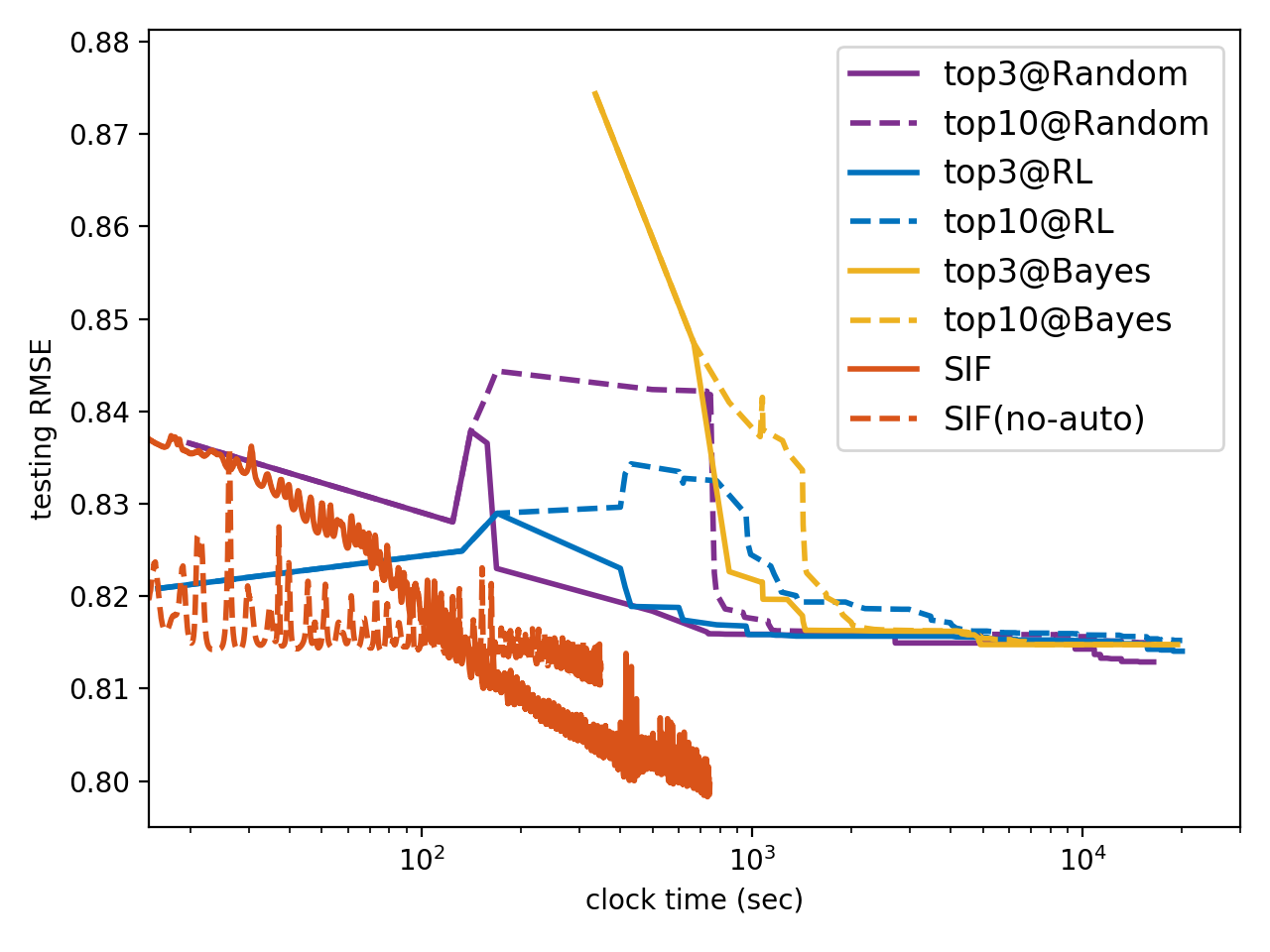}
		\label{fig:automc:1m}}
	\quad
	\subfigure[Youtube.]
	{\includegraphics[width = 0.29\textwidth]{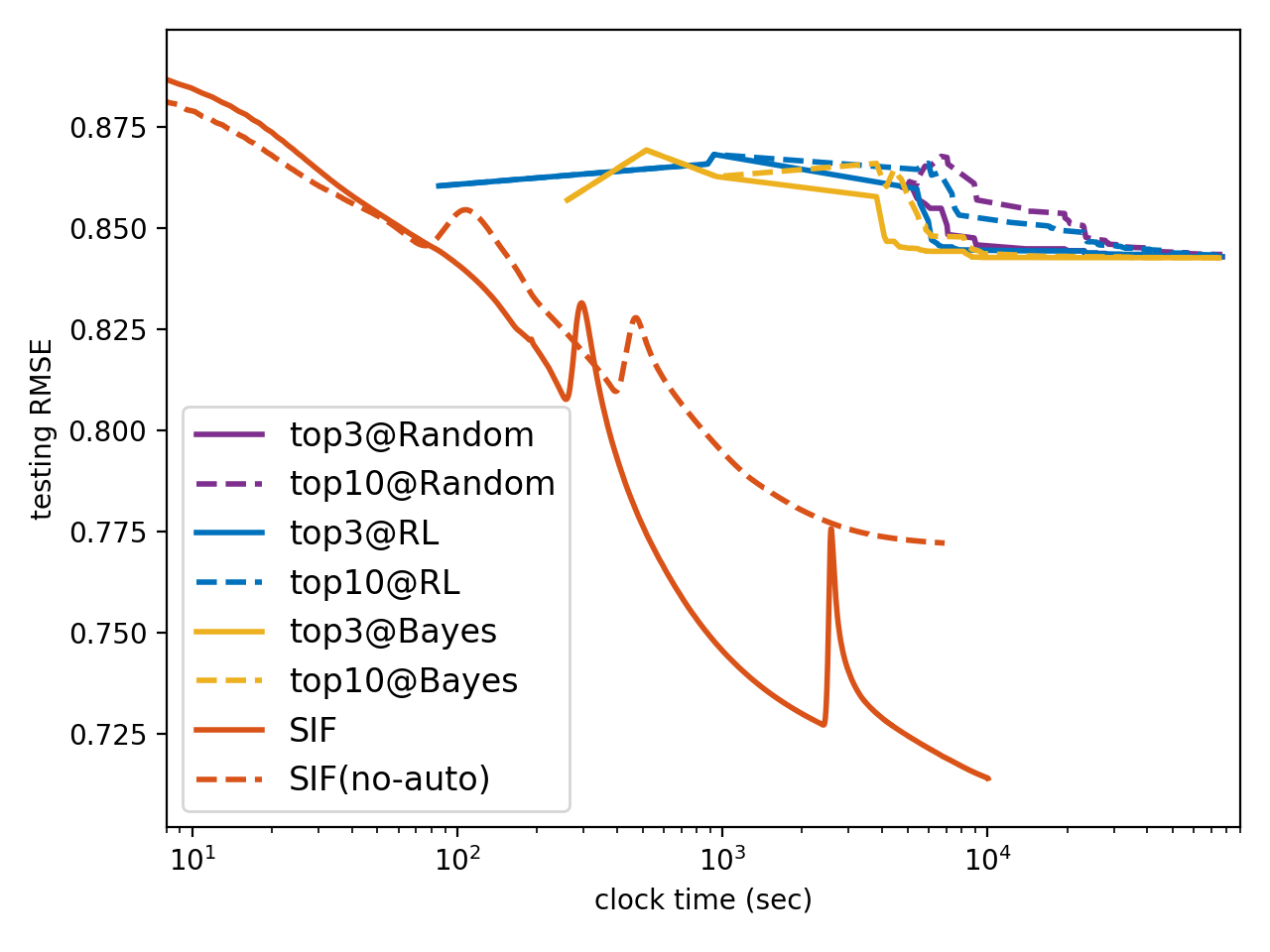}
		\label{fig:automc:ysub}}
	
	\vspace{-14px}
	\caption{Search efficiency of \textit{SIF} and the other AutoML approaches
		(embedding dimension is $8$).}
	\label{fig:automc:speed}
	\vspace{-5px}
\end{figure*}

\subsubsection{More Performance Metrics}
As in \cite{kim2016convolutional,NeuMF2017,he2018outer},
we report the metrics of ``Hit at top" and ``Normalized Discounted Cumulative Gain (NDCG) at
top" on the
MovieLens-100K data.
Recall that the ratings are in the range $\{1, 2, 3 ,4, 5\}$.
We treat ratings that are equal to five as positive, and the others as negative.
Results are shown in Table~\ref{tab:morem}.
The comparison between SIF and SIF(no-auto) shows that using the validation set can lead to better architectures.
Besides,
SIF is much better than the other methods in terms of both Hit@K and NDCG@K, and
the relative improvements are larger than that on RMSE.

\begin{table}[ht]
\caption{Hit-at-top (H@K) and NDCG-at-top (N@K) on MovieLens-100K.}
\vspace{-10px}
\begin{tabular}{c | c | c | c | c | c}
	\hline
	                     & RMSE  & H@5   & H@10  & N@5   & N@10  \\ \hline
	      Altgrad        & 0.867 & 0.267 & 0.377 & 0.156 & 0.220 \\ \hline
	         FM          & 0.845 & 0.286 & 0.391 & 0.176 & 0.249 \\ \hline
	 $\!$Deep\&Wide$\!$  & 0.861 & 0.273 & 0.378 & 0.163 & 0.227 \\ \hline
	        NCF          & 0.851 & 0.279 & 0.386 & 0.172 & 0.236 \\ \hline
	$\!$SIF(no-auto)$\!$ & 0.846 & 0.284 & 0.390 & 0.175 & 0.250 \\ \hline
	        SIF          & \textbf{0.839} & \textbf{0.295} & \textbf{0.405} & \textbf{0.190} & \textbf{0.259} \\ \hline
\end{tabular}
\label{tab:morem}
\end{table}

\subsection{Comparison with State-of-the-Art AutoML Search Algorithms}
\label{ssec:exp:auto}

In this section, we compare with
the following popular AutoML approaches: 
(i) ``\textit{Random}'':
Random search \citep{bergstra2012random} is used.
Both operations and weights (for the small and fixed MLP) 
in the designed search space (in Section~\ref{sec:space}) are uniformly and
randomly set;
(ii) 
``\textit{RL}'':
Following \citep{zoph2017neural},
we use reinforcement learning \citep{sutton1998reinforcement}
to search the designed space;
(iii) ``\textit{Bayes}'':
The designed search space is optimized by HyperOpt \citep{bergstra2015hyperopt},
a popular Bayesian optimization approach for hyperparameter tuning; and
(iv) 
``\textit{SIF}'':
The proposed Algorithm~\ref{alg:sif};
and (v) 
``\textit{SIF(no-auto)}'':
A variant of 
\textit{SIF}
in which parameter $\bm{S}$ for the IFCs are also optimized with training data.
More details on the implementations and discussion of the other AutoML approaches are in Appendix~\ref{app:details:auto}.


\subsubsection{Effectiveness}

Figure~\ref{fig:auto}
shows
the testing RMSEs of the various AutoML approaches.
Experiments on MovieLens-10M are not performed as the other baseline methods 
are very slow
(Figure~\ref{fig:automc:speed}).
\textit{SIF(no-auto)} is worse than \textit{SIF} as the
IFCs are searched purely based on the training set.
Among all the methods tested,
the proposed \textit{SIF} is the best.
It can find good IFCs, leading to lower testing RMSEs than the other methods
for the various embedding dimensions.

\begin{figure*}[ht]
	\centering
	\subfigure[Operations (vector-wise).]
	{\includegraphics[width = 0.32\textwidth]{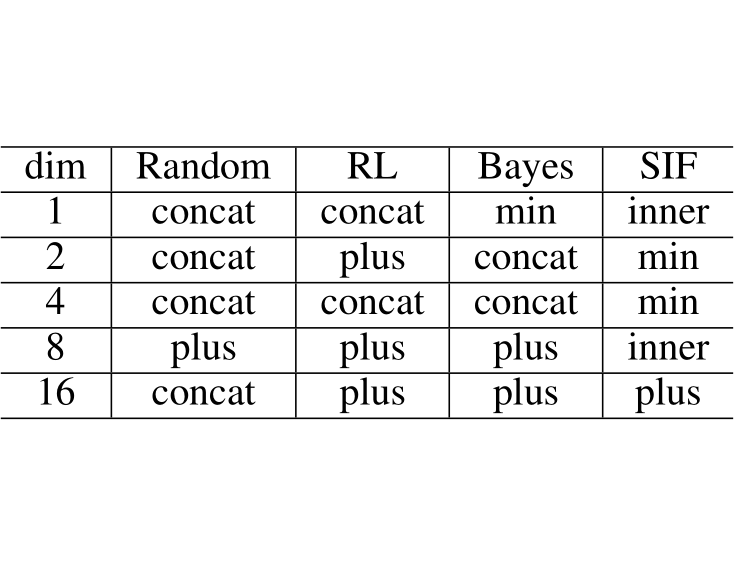}
		\label{tab:seops}}
	\quad
	\subfigure[Nonlinear transformation (element-wise).]
	{\includegraphics[width = 0.25\textwidth]{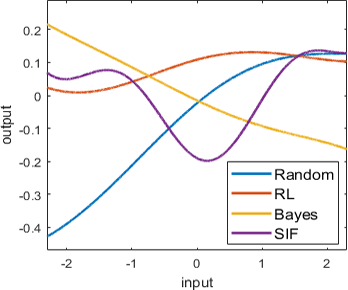}
		\label{fig:eletrans}}
	\quad
	\subfigure[Performance of each single operation.]
	{\includegraphics[width = 0.26\textwidth]{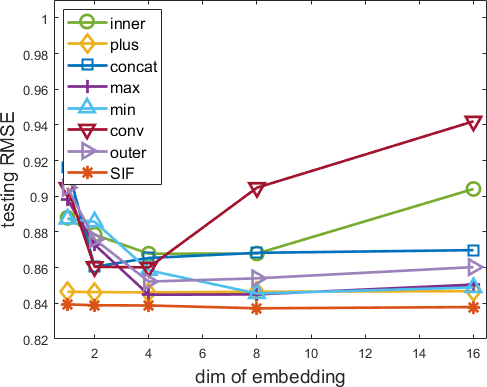}
		\label{fig:sing}}
	
	\vspace{-14px}
	\caption{(a) operations identified by various search algorithms on MovienLens-100K;
		(b) Searched IFCs on MovienLens-100K when embedding dimension is $8$;
		(c) Performance for SIF and each single operation on MovieLens-100K.}
	\vspace{-10px}
\end{figure*}

\begin{figure*}[ht]
	\centering
	\subfigure[embedding dimension = 4.]
	{\includegraphics[width = 0.29\textwidth]{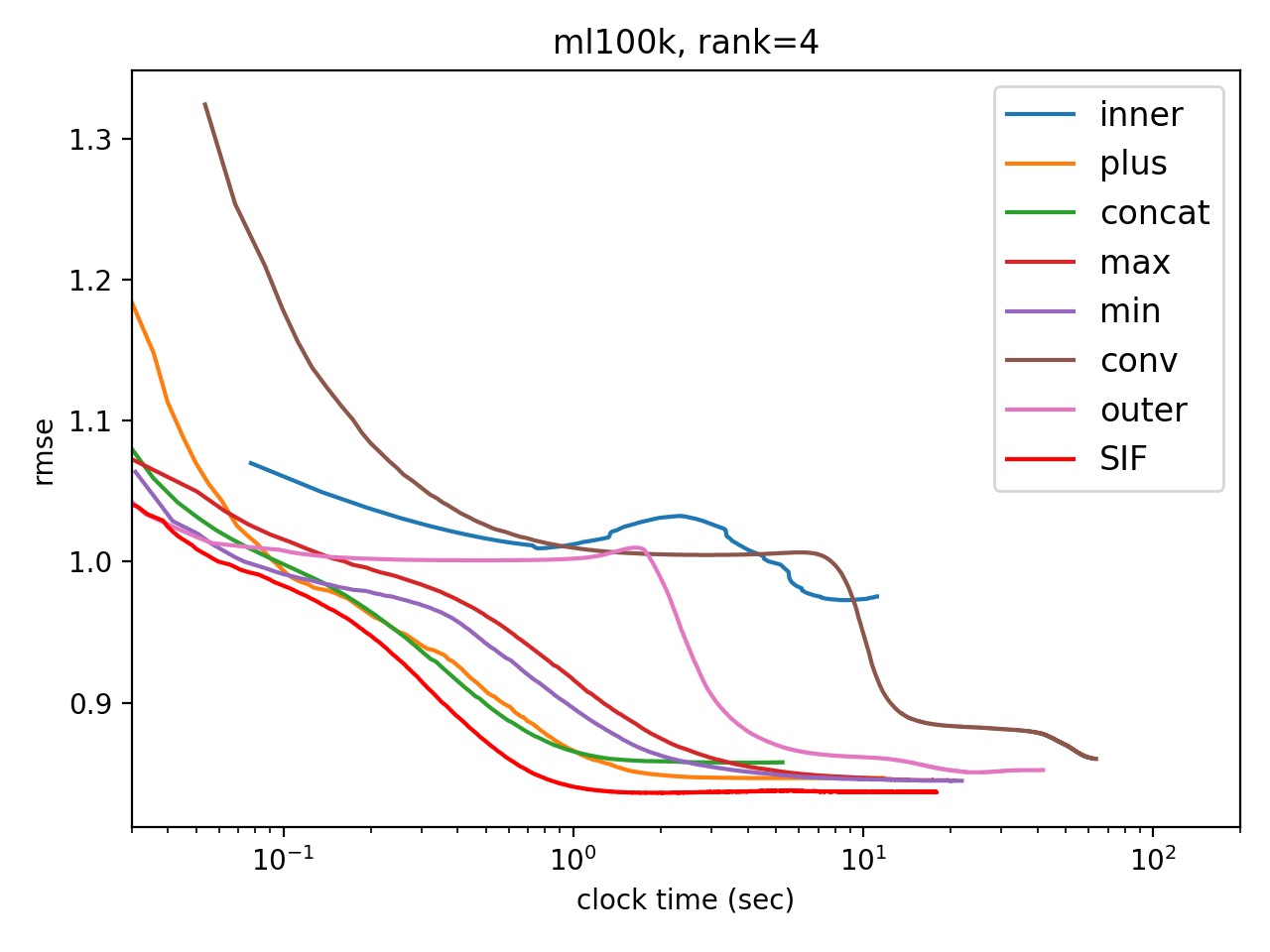}}
	\quad
	\subfigure[embedding dimension = 8.]
	{\includegraphics[width = 0.29\textwidth]{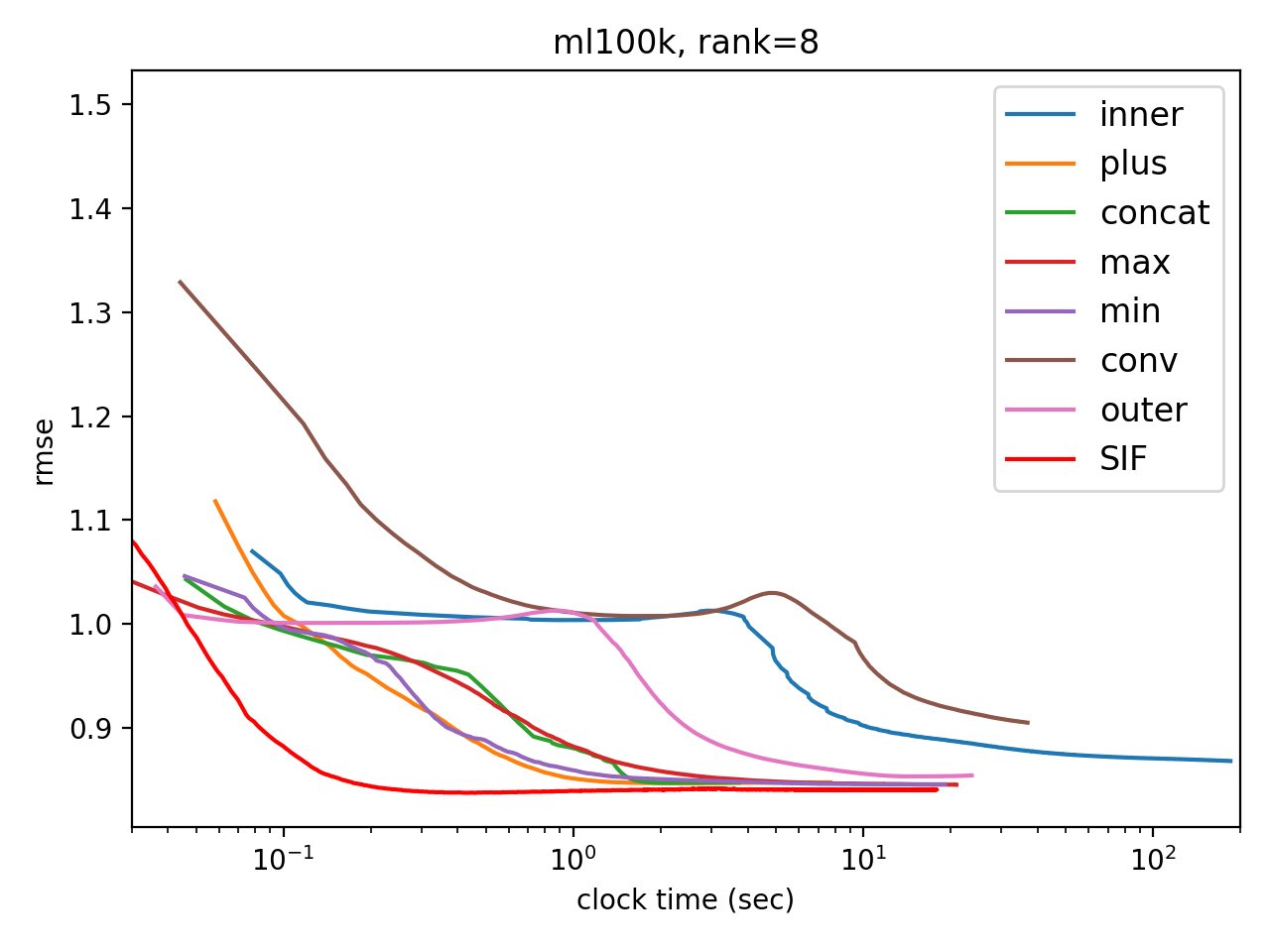}}
	\quad
	\subfigure[embedding dimension = 16.]
	{\includegraphics[width = 0.29\textwidth]{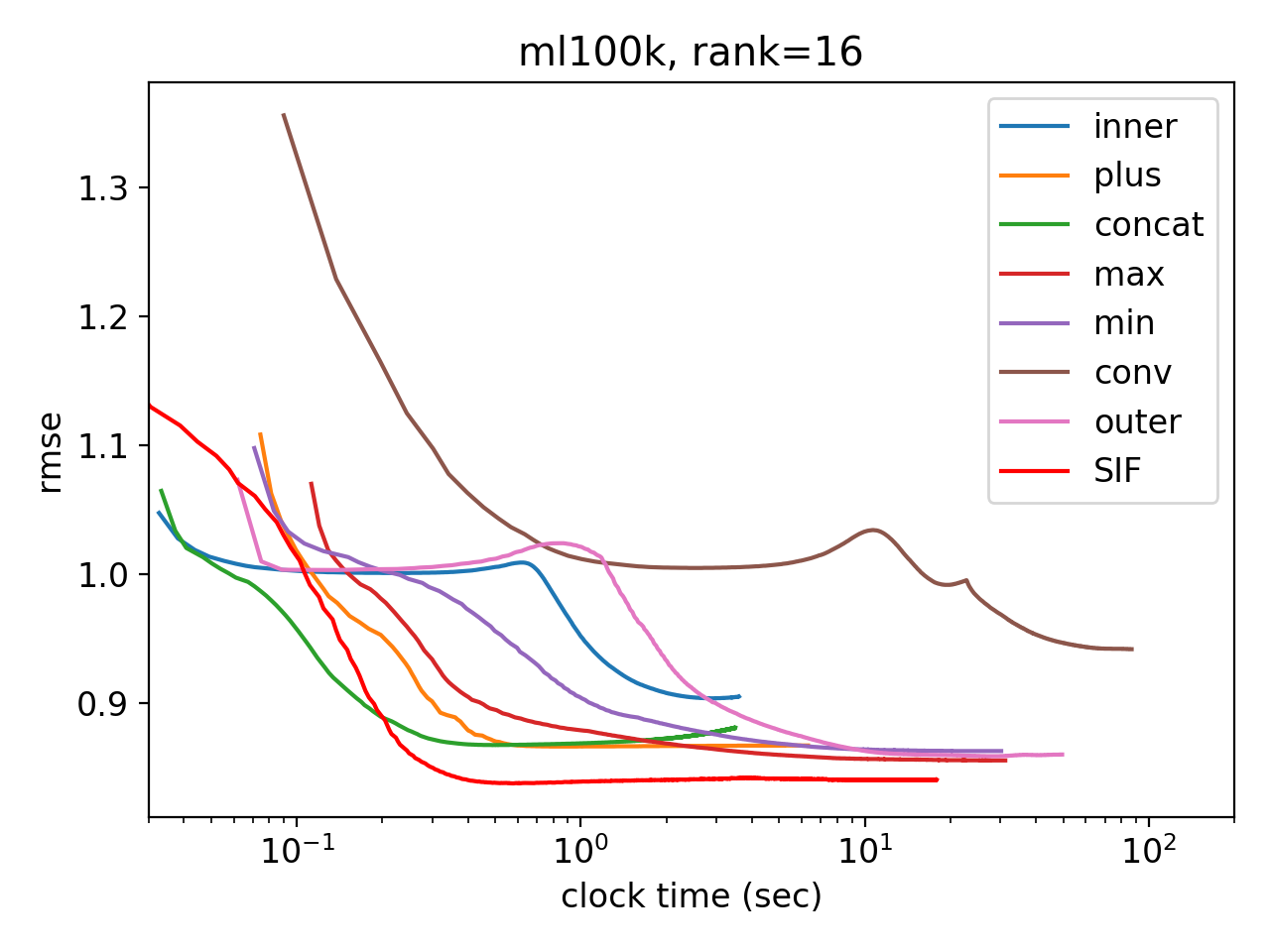}}
	
	\vspace{-14px}
	\caption{Convergence of various single operations on MovieLens-100K, with
	different embedding dimensions.}
	\label{fig:abla:ml100k}
	\vspace{-5px}
\end{figure*}

\subsubsection{Search Efficiency}

In this section, we take
the $k$ 
architectures with top
validation performance, 
re-train, and then
report
their average
RMSE
on the testing set  in
Figure~\ref{fig:automc:speed}.
As can be seen, 
all algorithms run slower on Youtube,
as the search space for tensor data is larger than that for matrix data.
Besides,
\textit{SIF} is much faster than all the other methods and has
lower testing RMSEs.
The gap is larger on the Youtube data set.
Finally,
Table~\ref{tab:ftime}
reports the time spent on the search and fine-tuning.
As can be seen,
the time taken by SIF is less than five times of those of the other non-autoML-based methods.


\begin{table*}[ht]
\caption{Clock time (in seconds) taken by SIF and the other CF approaches  
(embedding dimension is $8$).}
	\vspace{-10px}
	\centering
	\begin{tabular}{c | C{35px} | C{35px} | c | C{35px} || C{35px} | c}
		\hline
		                           & AltGrad & FM    & Deep\&Wide & NCF   & SIF   & SIF(no-auto) \\ \hline
		MovieLens-100K & 25.4    & 43.1  & 37.9       & 34.3  & 159.8 & 73.4         \\ \hline
		                           MovieLens-1M   & 313.7   & 324.3 & 357.0      & 374.9 & 745.3 & 348.7        \\ \hline
	\end{tabular}
	
%
	\label{tab:ftime}
	
\end{table*}

\begin{figure*}[ht]
	\centering
	\subfigure[MovieLens-100K.]
	{\includegraphics[width = 0.29\textwidth]{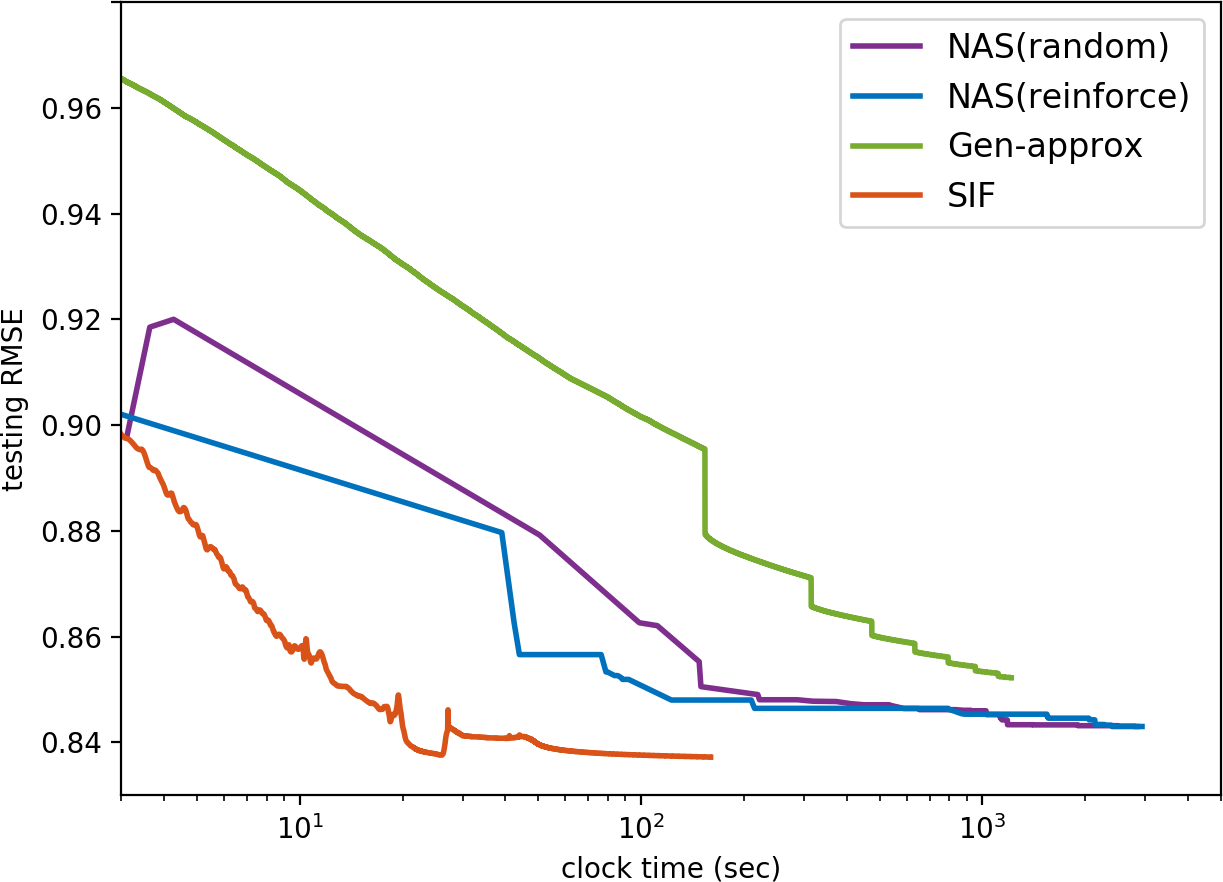}}
	\quad
	\subfigure[MovieLens-1M.]
	{\includegraphics[width = 0.29\textwidth]{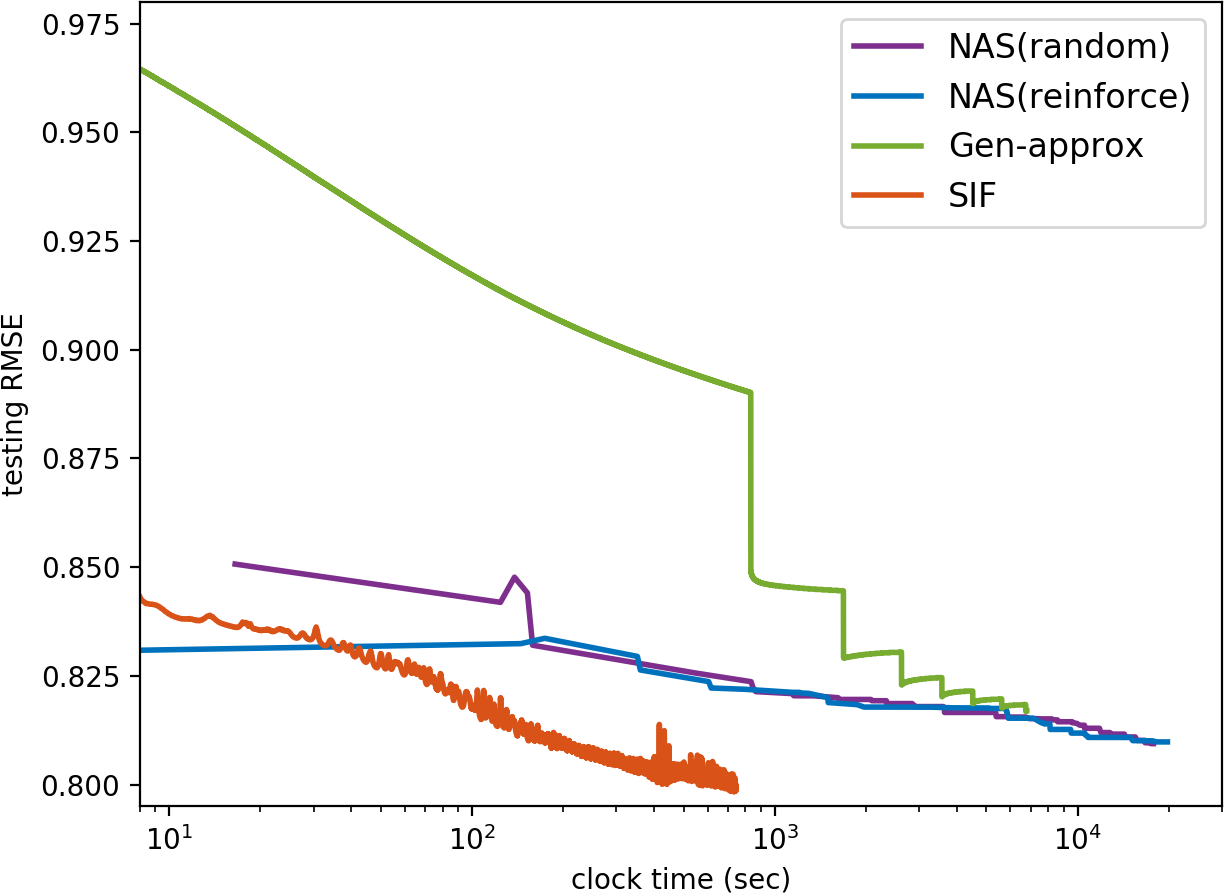}}
	\quad
	\subfigure[Youtube.]
	{\includegraphics[width = 0.29\textwidth]{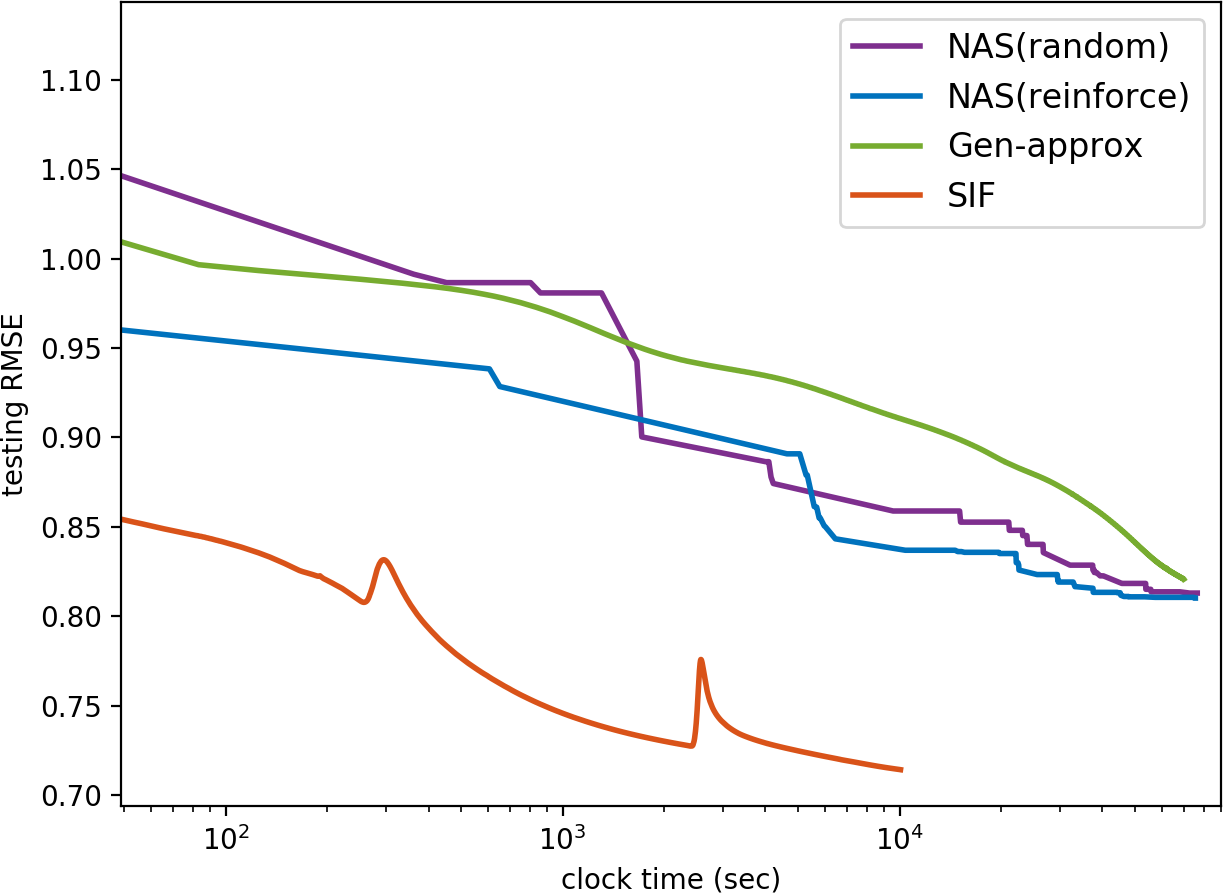}}
	
	\vspace{-14px}
	\caption{Comparison on different search space designs  (embedding dimension is
	8).}
	\label{fig:abla:space}
	\vspace{-5px}
\end{figure*}

\subsection{Interaction Functions (IFCs) Obtained}
\label{sec:exp:single}

To understand why a lower RMSE can be achieved by the proposed method,
we show the IFCs obtained by the various AutoML methods on
MovieLens-100K.
Figure~\ref{tab:seops}
shows the vector-wise operations obtained.
As can be seen, \textit{Random}, \textit{RL}, \textit{Bayes} and \textit{SIF}
select different operations
in general.
Figure~\ref{fig:eletrans}
shows
the searched nonlinear transformation for each element.
We can see that \textit{SIF} can find more complex transformations than the others.

To further demonstrate the need of AutoML and effectiveness of SIF,
we show the performance of each single operation in
Figure~\ref{fig:sing}.
It can be seen that 
while some operations can be better than others
(e.g., {\sf plus} is better than {\sf conv}),
there is no clear winner among all operations.
The best operation may depend on the embedding dimension as well.
These verify the need for AutoML.
Figure~\ref{fig:abla:ml100k} shows the testing RMSEs of all single operations.
We can see that SIF consistently achieves lower testing RMSEs than all single operations and converges faster.
Note that SIF in Figure~\ref{tab:seops} may not select the best single operation in Figure~\ref{fig:sing},
due to the learned nonlinear transformation 
(Figure~\ref{fig:eletrans}).

\begin{table*}[ht]
	\centering
	\caption{Results on 
		MovieLens-100K with
	$k$ selected operations in SIF.}
	\vspace{-10px}
	\begin{tabular}{c | C{35px}  C{120px} | C{35px}  C{120px}}
		\hline
		&                   \multicolumn{2}{c|}{embedding dimension = 4}                   &                   \multicolumn{2}{c}{embedding dimension = 8}                    \\ \cline{2-5}
		$k$ & RMSE            & operator                                                       & RMSE            & operator                                                       \\ \hline
		1        & 0.8448          & {\sf concat}                                                   & 0.8450          & {\sf max}                                                      \\ \hline
		2        & 0.8435          & {\sf concat}, {\sf max}                                        & 0.8440          & {\sf max}, {\sf plus}                                          \\ \hline
		3        & 0.8442          & {\sf concat}, {\sf max}, {\sf multiply}                        & 0.8432          & {\sf max}, {\sf plus}, {\sf concat}                            \\ \hline
		4        & 0.8433          & {\sf concat}, {\sf max}, {\sf multiply}, {\sf plus}            & 0.8437          & {\sf max}, {\sf plus}, {\sf concat}, {\sf min}                 \\ \hline
		5        & \textbf{0.8432} & {\sf concat}, {\sf max}, {\sf multiply}, {\sf plus}, {\sf min} & \textbf{0.8431} & {\sf max}, {\sf plus}, {\sf concat}, {\sf min}, {\sf multiply} \\ \hline
	\end{tabular}
	\label{tab:moreops}
\end{table*}

\begin{table*}[ht]
\caption{Testing RMSE 
on MovieLens-100K 
with different activation functions and number of hidden units 
in the MLP for element-wise transformation.}
	\centering
	\vspace{-10px}
	\begin{tabular}{c|C{40px}|C{35px}|C{35px}|C{35px}|C{35px}|C{35px}}
		\hline
		embedding & activation &     \multicolumn{5}{c}{number of hidden units}      \\ \cline{3-7}
		dimension &  function  &   1    &   5    &       10        &   15   &   20   \\ \hline
		&    relu    & 0.8437 & 0.8388 & \textbf{0.8385} & 0.8389 & 0.8396 \\ \cline{2-7}
		4     &  sigmoid   & 0.8440 & 0.8391 &     0.8390      & 0.8395 & 0.8399 \\ \cline{2-7}
		&    tanh    & 0.8439 & 0.8991 &     0.8389      & 0.8393 & 0.8401 \\ \hline\hline
		&    relu    & 0.8385 & 0.8372 & \textbf{0.8370} & 0.8371 & 0.8374 \\ \cline{2-7}
		8     &  sigmoid   & 0.8382 & 0.8375 &     0.8377      & 0.8376 & 0.8378 \\ \cline{2-7}
		&    tanh    & 0.8386 & 0.8376 &     0.8373      & 0.8375 & 0.8377 \\ \hline
	\end{tabular}
	\label{tab:element}
	\vspace{-5px}
\end{table*}

\subsection{Ablation Study}

In this section,
we perform ablation study on 
different parts of the proposed AutoML method.

\subsubsection{Different Search Spaces}
First,
we show the superiority
of search space used in SIF by comparing with
the following approaches:
\begin{itemize}[leftmargin = 9px]
\item Using a MLP as a general approximator (``\textit{Gen-approx}''), as described in Section~\ref{sec:space},
to approximate the search space is also compared.
The MLP is updated with stochastic gradient descent \citep{bengio2000gradient} using the validation set.
Since searching network architectures 
is expensive
\citep{zoph2017neural,zoph2017learning}, the MLP structure
is fixed for \textit{Gen-approx}
(see Appendix~\ref{app:mlpdetails}).

\item Standard NAS approach,
using MLP
to approximate the IFC $f$.
The MLP is optimized with the training data,
while its architecture is searched with the validation set.
Two search algorithms
are considered:
(i) random search (denoted `\textit{`NAS(random)}'') \cite{bergstra2012random};
(ii) reinforcement learning (denoted ``\textit{NAS(reinforce)}'') \cite{zoph2017neural}.
\end{itemize}
The above are general search spaces,
and are much larger than the one designed for SIF.

Figure~\ref{fig:abla:space}
shows the convergence of testing RMSE for the various methods. 
As can be seen, these general approximation methods 
are hard to be searched and thus much slower than SIF.
The proposed search space in Section~\ref{sec:space}
is not only compact,
but also allows
efficient one-shot search as discussed in Section~\ref{sec:algorithm}.

\subsubsection{Allowing More Operations}
In Algorithm~\ref{alg:sif},
we only allow one operation to be selected.
Here,
we allow 
more operations by changing
$\mathcal{C}_1$ to $\mathcal{C}_k = \left\lbrace \bm{\alpha} \,|\, \NM{\bm{\alpha}}{0} = k \right\rbrace$,
where $k \in \{ 1, 2,\dots, 5 \}$.
Results are shown in Table~\ref{tab:moreops}.
As can be seen,
the testing RMSE
gets slightly 
smaller.
However,
the model complexity and prediction time
grow linearly with $k$, and so 
can become significantly larger.

\subsubsection{Element-wise Transformation}
Recall that in Section~\ref{sec:space},
we use a small MLP to approximate an arbitrary element-wise transformation.
In this experiment, we vary the number of hidden units and type of activation function in
this MLP.
Results on the testing RMSE are shown in Table~\ref{tab:element}.
As can be seen,
once the number of hidden units is large enough (i.e., $\ge 5$ here),
the performance is stable
with different number of activation functions.
This
demonstrates the robustness of our design in the search space.

\subsubsection{Changing Predictor to MLP}
In \eqref{eq:gencf},
we used a linear predictor.
Here, 
we study whether using a more complicated predictor can further boost learning performance.
A standard three-layer MLP with 10 hidden units is used.
Results are shown in Table~\ref{tab:chpred}.
As can  be seen,
using a more complex predictor
can 
lead to lower testing RMSE when the embedding dimension is 4, 8, and 16.
However,
the lowest testing RMSE is still achieved by the linear predictor with an embedding
dimension of 2.
This demonstrates that
the proposed SIF can achieve the desired performance,
and designing a proper predictor is not an easy task.

\begin{table}[ht]
	\centering
	\caption{Testing RMSE  on 
		MovieLens-100K 
	with MLP and linear predictor in \eqref{eq:gencf}.}
	\vspace{-10px}
	\begin{tabular}{c|c|c|c|c}
		\hline
		    &     \multicolumn{2}{c|}{MLP}     & \multicolumn{2}{c}{linear} \\ \hline
		embedding dim &      RMSE       &    operator    &  RMSE  &     operator      \\ \hline
		 2  &     0.8437      &  {\sf concat}  & \textbf{0.8389} &     {\sf min}     \\ \hline
		 4  &     0.8424      &  {\sf concat}  & 0.8429 &     {\sf min}     \\ \hline
		 8  & 0.8407 &   {\sf plus}   & 0.8468 &    {\sf inner}    \\ \hline
		16  &     0.8413      & {\sf multiply} & 0.8467 &    {\sf plus}     \\ \hline
	\end{tabular}
	\label{tab:chpred}
	\vspace{-10px}
\end{table}

\section{Conclusion}

In this paper,
we propose an AutoML approach to search for interaction functions
in CF.
The keys  for its success
are (i) an expressive search space, (ii)
a continuous representation of the space, 
and (iii) an efficient algorithm
which can jointly search interaction functions and update embedding vectors
in a stochastic manner.
Experimental results demonstrate that
the proposed method 
is much more efficient than popular AutoML approaches,
and also obtains much better learning performance
than human-designed CF approaches.

\appendix

\section{Appendix}

\subsection{General Search Space}
\label{app:mlpdetails}

As in Figure~\ref{fig:space} and \eqref{eq:space},
we can take a three-layer MLP as $\mathcal{F}$,
which is guaranteed to approximate any given function with enough hidden units \cite{raghu2017expressive}.
We concatenate $\bm{u}_i$ and $\bm{v}_j$ as input to MLP.
To ensure the approximation ability of MLP,
we set the number of hidden units to be double that of the input size,
and use the sigmoid function as the activation function.
The final output is a vector
of the same dimension as $\bm{u}_i$.

\subsection{Tensor Data}
\label{app:tensor}

\begin{figure}[ht]
	\centering
	\includegraphics[width=0.425\textwidth]
	{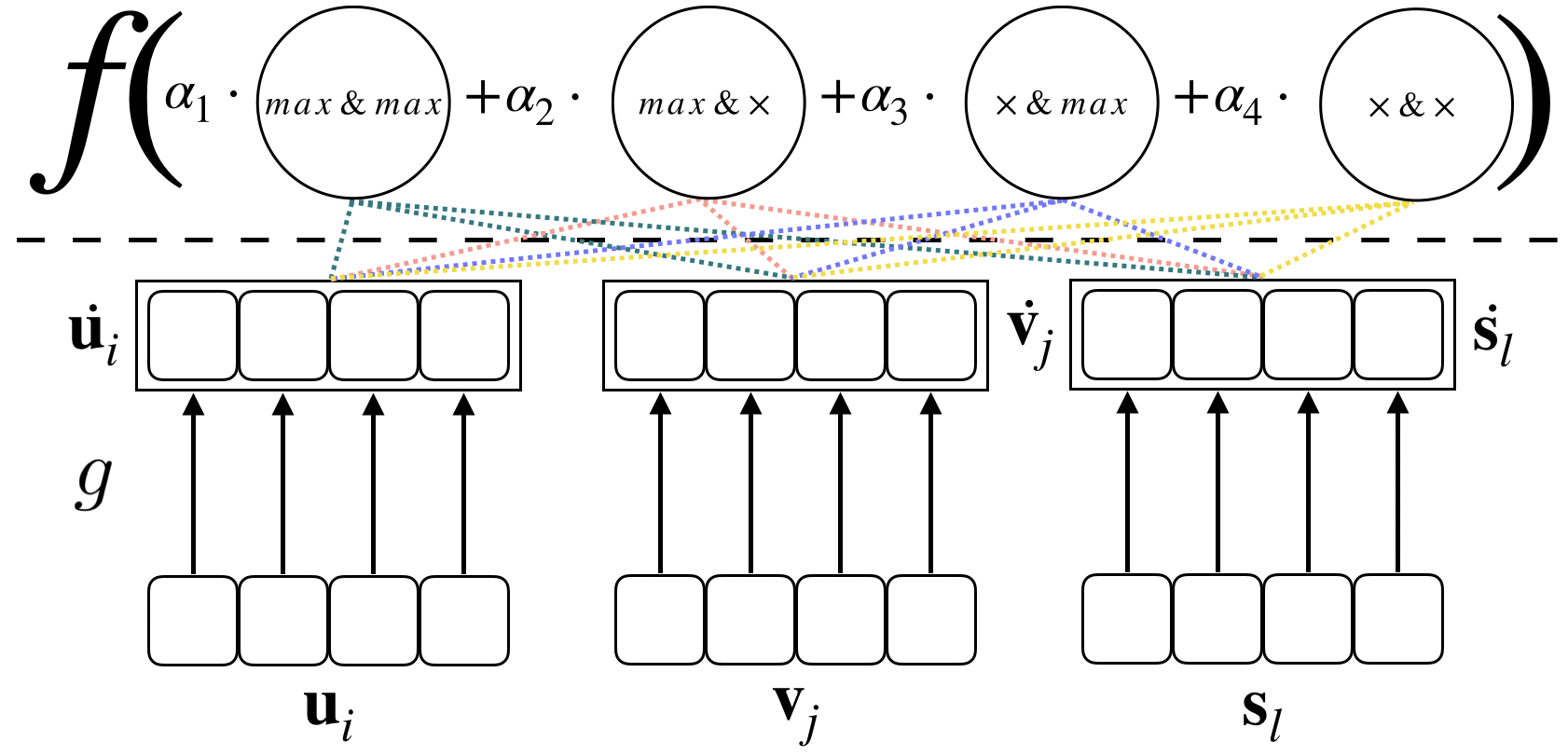}
	\vspace{-10px}
	\caption{Representation of the search space for tensor data.}
	\label{fig:tensor}
	\vspace{-10px}
\end{figure}

Following Section~\ref{sec:tensor},
the proposed Algorithm~\ref{alg:sif} can be extended to tensor data.
If only the times and max operations are allowed,
the search space can be represented as in Figure~\ref{fig:tensor},
which is similar to Figure~\ref{fig:space}.
Note that two possible operations are chosen, 
so the search space for tensor data is much larger than that for matrix data
($O(K)$ vs $O(K^2)$, where $K$ is the number of operations).

\subsection{Proofs of Proposition~\ref{pr:consw}}
\label{app:pr:consw}

\begin{proof}
	Taking $f$ as the inner product function as an example.
	Let $\bm{A} = \{ \bm{U}, \bm{V}, \bm{w} \} \neq \bm{0}$
	be an optimal point of $F$,
	then
	$F(\bm{A}) = 
	\sum\nolimits_{(i,j)\in\Omega}
	\ell( \bm{w}^{\top} f ( \bm{u}_i, \bm{v}_j ), \bm{O}_{ij}  )^2
	+ \nicefrac{\lambda}{2}\NM{\bm{U}}{F}^2
	+ \nicefrac{\lambda}{2}\NM{\bm{V}}{F}^2$.
	We construct another $\bm{A}' = \{ \beta \bm{U}, \beta \bm{V}, \nicefrac{1}{\beta^2} \bm{w} \}$
	with $\beta \in (0, 1)$,
	then 
	$	F\left( \bm{A}'\right) 
	= \sum_{(i,j)\in\Omega} 
	\ell( \bm{w}^{\top} f ( \bm{u}_i, \bm{v}_j ), \bm{O}_{ij}  )^2
	+ \nicefrac{\lambda \beta^2}{2}\NM{\bm{U}}{F}^2
	+ \nicefrac{\lambda \beta^2}{2}\NM{\bm{V}}{F}^2
	< F(\bm{A})$,
	which violates the assumption that $\bm{A} \neq \bm{0}$ is an optimal solution.
	The same holds for $f$ being other operations in Table\ref{tab:operations}.
\end{proof}

\subsection{Implementation: CF Approaches}
\label{app:details:cf}

\noindent
\textbf{AltGrad}:
It is the traditional way to perform collaborative filtering. 
We first apply an element-wise product of the user and item embedding and then feed the outputs into a linear predictor. 
In other word, AltGrad is equivalent to using the single inner operation in our searching space.

\noindent
\textbf{Factorization Machine}:
For the matrix case, we directly utilize the implementation from pyFM \footnote{\url{https://github.com/coreylynch/pyFM}}, 
noting that pyFM is difficult to run on GPU so it is not comparable to other methods when it comes to training time. 
For tensor case (HOFM), we use the implementation from tffm \footnote{\url{https://github.com/geffy/tffm}}, 
which can be easily accelerated by GPU since tffm is implemented with tensorflow.

\noindent
\textbf{Deep \& Wide}:
We implement Deep \& Wide by employing a two layer MLP with ReLU as the non-linear function on the concatenation of all potential embeddings.

\noindent
\textbf{NCF}:
Neural collaborative filtering is flexible to stack many layers and become very deep as well as learn separate embeddings for GMF and MLP. 
But this paper focuses on the interaction function, and also to ensure similar computational complexity, 
we implement NCF by combining generalized matrix factorization (GMF) with a one-layer multi-layer perceptron (MLP). 
Noting that our method also supports deep models by changing the last linear predictor to a deep MLP.

\noindent
\textbf{CP}:
Similar to AltGrad, CP first combines three embeddings by an element-wise product, then the prediction is carried out through a linear predictor.

\noindent
\textbf{Tucker}:
Tucker has high computation complexity since it has a 3-D weight to perform Tucker decomposition. 
We implement this method by sequentially applying tensor product along all three dimensions and then feed the result into a linear predictor.

\noindent
\textbf{Others}:
Note that,
CML \cite{hsieh2017collaborative,Zhang2016CollaborativeKB}, 
ConvMF \cite{kim2016convolutional} and ConvNCF \cite{he2018outer} are not included,
since their CF tasks are different and codes are not available.
Instead, 
interaction functions they introduced (Table~\ref{tab:operations}) are studied in Section~\ref{sec:exp:single}.

\subsection{Implementation: AutoML Approaches}
\label{app:details:auto}

\noindent
\textbf{General Approximator}:
As in AutoML literature,
the search space needs to be carefully designed,
it cannot be too large (hard to be searched) nor too small (poor performance).
In the experiments,
we use MLP structure in Appendix~\ref{app:mlpdetails}.
The standard approach to optimize MLP is gradient descent on hyper-parameters
(please see \cite{bergstra2012random}).
However,
it is very slow ---
in order to perform one gradient descent on MLP,
we need to finish the training of CF model,
which is one full model training on the training dataset.
MLP needs many iterations to converge,
and thus to train a good MLP,
we need many times of full model training.
This makes Gen-Approx very slow.

\noindent
\textbf{Random}:
In this baseline, architecture is randomly generated including the weights $\bm{p}$, 
$\bm{q}$ for element-wise MLP and the interaction function $f$ in every epoch. Every weights of $\bm{p}$, $\bm{q}$ is restricted within $[-3.0,3.0]$. 
We then report the best RMSE
achieved after a fixed number of architectures are sampled.

\noindent
\textbf{Bayes}:
We directly use the source code from hyperopt \cite{bergstra2015hyperopt}
to perform Bayesian optimization. 
Every single weight of $\bm{p}$, $\bm{q}$ is designed as a uniform space ranging from -3.0 to 3.0. 
This continuous space along with the discrete space of interaction function is then jointly optimized, we report the best RMSE until a fixed number of evaluations are achieved.

\noindent
\textbf{Reinforcement Learning}:
Following \cite{zoph2017neural},
we utilize a controller to generate the architecture including $\bm{p}$, $\bm{q}$ and the interaction function $f$ to combine all potential embeddings. The difference lies in that the searching space in our setting is a combined continuous ($\bm{p}$, $\bm{q}$) and discrete (interaction function) space, so deterministic policy gradient is utilized \cite{Silver2014DPG, Lillicrap2015ContinuousCW} to train the controller. We employ a recurrent neural network to act as the controller in order to be flexible to both matrix and tensor. 
The controller is then trained via policy gradient where the reward is $\nicefrac{1}{\text{RMSE}}$ of the generated architecture on the validation set. 
At convergence, a neural network is built following the output of the controller and the RMSE on test set is recorded.

\noindent
\textbf{Others}:
SMAC \cite{feurer2015efficient} is not compared as it cannot be run on our GPU cluster
and HyperOpt has comparable performance \cite{kandasamy2019tuning}.
Genetic algorithms \cite{automl_book}
are also not compared as they need special designs to fit into our search space,
and is inferior to RL and random search \cite{zoph2017neural,liu2018darts}.
	

%
%
%
%
%
%
%
%



\clearpage
\bibliographystyle{ACM-Reference-Format}
\bibliography{bib}

\end{document}